\documentclass[lettersize,journal]{IEEEtran}

\usepackage{amsmath,amsfonts}
\usepackage{algorithm}
\usepackage{algpseudocode}
\usepackage{array}
\usepackage[caption=false,font=normalsize,labelfont=sf,textfont=sf]{subfig}
\usepackage{textcomp}
\usepackage{stfloats}
\usepackage{url}
\usepackage{verbatim}
\usepackage{graphicx}
\usepackage{pgf}
\usepackage{layouts}
\usepackage{csquotes}
\usepackage{subfiles}

\def\BibTeX{{\rm B\kern-.05em{\sc i\kern-.025em b}\kern-.08em
		T\kern-.1667em\lower.7ex\hbox{E}\kern-.125emX}}
\usepackage{balance}

\usepackage{cite}
\usepackage{bm}
\usepackage{mathtools}
\usepackage{siunitx}
\usepackage{import} %
\usepackage[hidelinks]{hyperref}
\usepackage{booktabs}
\usepackage{multirow}
\usepackage{tabularx}
\usepackage{xr-hyper} %
\usepackage{tikz}
\usepackage{lipsum}

\usepackage{amsthm}

\newtheorem{assumption}{Assumption}
\newtheorem{lemma}{Lemma}
\newtheorem{theorem}{Theorem}

\newtheorem{res}{Result}
\newtheorem{defn}{Definition}

\newcommand\textpreprint{%
	\footnotesize This work has been published under a Creative Commons Attribution 4.0 License
	at IEEE Access with Digital Object Identifier \href{https://doi.org/10.1109/ACCESS.2023.3329685}{10.1109/ACCESS.2023.3329685}
}

\newcommand\preprintnotice{%
	\begin{tikzpicture}[remember picture,overlay]
		\node[anchor=south,yshift=10pt] at (current page.south) {\fbox{\parbox{\dimexpr\textwidth-\fboxsep-\fboxrule\relax}{\textpreprint}}};
	\end{tikzpicture}%
}
\usepackage{amsmath}
\usepackage{amsfonts}
\usepackage{amssymb}
\usepackage{bm}

\newcommand{\NN}{\scriptscriptstyle \text{NN}}
\newcommand{\test}{{\scriptscriptstyle \mathrm{test}}}
\newcommand{\train}{{\scriptscriptstyle \mathrm{train}}}
\newcommand{\val}{{\scriptscriptstyle \mathrm{val}}}

\def\DKL{{D_\text{KL}}}
\def\vec{\mathrm{vec}}
\def\diag{\mathrm{diag}}

\def\rt{{\textnormal{t}}}

\def\rvepsilon{{\mathbf{\epsilon}}}

\def\rvt{{\mathbf{t}}}

\def\rvw{{\mathbf{w}}}

\def\rvy{{\mathbf{y}}}

\def\vzero{{\bm{0}}}
\def\vone{{\bm{1}}}
\def\valpha{{\bm{\alpha}}}
\def\vmu{{\bm{\mu}}}
\def\vnu{{\bm{\nu}}}

\def\vphi{{\bm{\phi}}}

\def\vepsilon{{\bm{\epsilon}}}
\def\vsigma{{\bm{\sigma}}}
\def\va{{\bm{a}}}

\def\ve{{\bm{e}}}

\def\vh{{\bm{h}}}

\def\vp{{\bm{p}}}

\def\vt{{\bm{t}}}

\def\vw{{\bm{w}}}
\def\vx{{\bm{x}}}
\def\vy{{\bm{y}}}

\def\mA{{\bm{A}}}
\def\mB{{\bm{B}}}

\def\mD{{\bm{D}}}
\def\mE{{\bm{E}}}

\def\mI{{\bm{I}}}

\def\mM{{\bm{M}}}

\def\mT{{\bm{T}}}

\def\mW{{\bm{W}}}
\def\mX{{\bm{X}}}

\def\mOne{{\bm{1}}}

\def\mPhi{{\bm{\Phi}}}
\def\mLambda{{\bm{\Lambda}}}
\def\mSigma{{\bm{\Sigma}}}

\def\gD{{\mathcal{D}}}

\def\gN{{\mathcal{N}}}

\def\sA{{\mathbb{A}}}

\def\sC{{\mathbb{C}}}

\def\sI{{\mathbb{I}}}

\def\sL{{\mathbb{L}}}

\def\sR{{\mathbb{R}}}
\def\sS{{\mathbb{S}}}

\def\sW{{\mathbb{W}}}
\def\sX{{\mathbb{X}}}

\def\sTheta{{\bm{\Theta}}}

\newcommand{\N}{\mathcal{N}}
\newcommand{\E}{\mathbb{E}}

\newcommand{\minimize}[1]{\underset{#1}{\mathrm{minimize}}}

\begin{document}
\title{
Improved uncertainty quantification for neural networks with Bayesian last layer
}
\author{Felix Fiedler and Sergio Lucia
\thanks{
	Felix Fiedler (\href{mailto:felix.fiedler@tu-dortmund.de}{felix.fiedler@tu-dortmund.de})
	and
	Sergio Lucia (\href{mailto:sergio.lucia@tu-dortmund.de}{sergio.lucia@tu-dortmund.de})
	are with the \href{https://pas.bci.tu-dortmund.de}{Chair of Process Automation Systems}
	at the TU Dortmund University,
	Emil-Figge-Straße 70, 44227 Dortmund, Germany.}
}

\markboth{}%
{}

\maketitle
\preprintnotice

\begin{abstract}
Uncertainty quantification is an 
important 
task in machine learning - a task in which 
standard
neural networks (NNs) have traditionally not excelled. 
This can be a limitation for safety-critical applications, 
where uncertainty-aware methods like Gaussian processes 
or Bayesian linear regression are often preferred. 
Bayesian neural networks are an approach to address this limitation. 
They assume probability distributions for all parameters and yield distributed predictions. 
However, training and inference are typically intractable 
and approximations must be employed. 
A promising approximation 
is NNs with Bayesian last layer (BLL). 
They assume distributed weights only in the linear 
output 
layer 
and yield a normally distributed prediction. 
To approximate the intractable Bayesian neural network, 
point estimates of the distributed weights in all 
but the last layer should be obtained by maximizing the marginal likelihood. 
This has previously been challenging, 
as the marginal likelihood is expensive to evaluate in this setting. 

We present a reformulation of the log-marginal likelihood of a NN with BLL 
which allows for efficient training using backpropagation. 
Furthermore, we address the challenge of uncertainty quantification for extrapolation points.  
We provide a metric to quantify the degree of extrapolation 
and derive a method to improve the uncertainty quantification for these points. 
Our methods are derived for the multivariate case and demonstrated in a simulation study. 
In comparison to Bayesian linear regression with fixed features,
and a Bayesian neural network trained with variational inference,
our proposed method achieves the highest log-predictive density on test data.

\end{abstract}

\begin{IEEEkeywords}
Bayesian last layer, Bayesian neural network, uncertainty quantification
\end{IEEEkeywords}

\section{Introduction}
\label{sec:introduction}

% \PARstart{M}{achine} 
Machine
learning tries to capture 
patterns and trends 
through data.
Both, the data and the 
identified patterns 
are subject to uncertainty
\cite{jospinHandsonBayesianNeural2022, abdar_review_2021}. 
For many applications, especially those where machine learning is applied to safety critical tasks, 
it is imperative to 
quantify the uncertainty of the predictions.
An important example is learning-based~
control,
where probabilistic system models are identified from data and used for safe control
decisions~\cite{harrisonControlAdaptationMetalearning2018, mckinnonMetaLearningPaired2021, wabersichNonlinearLearningbasedModel2021, dongRobustDataDrivenIterative2021, mckinnonLearningProbabilisticModels2019, hewingCautiousModelPredictive2020}.

Bayesian linear regression (BLR)
\cite{harrisonControlAdaptationMetalearning2018, mckinnonMetaLearningPaired2021, wabersichNonlinearLearningbasedModel2021, dongRobustDataDrivenIterative2021, mckinnonLearningProbabilisticModels2019} 
and Gaussian processes (GPs) 
\cite{mckinnonLearningProbabilisticModels2019, hewingCautiousModelPredictive2020} 
are prominent methods for probabilistic system identification.
Both assume that a nonlinear feature space can be mapped linearly to the outputs.
As a main difference,
BLR requires the features to be explicitly defined,
while for GPs
the features are implicitly defined through 
a kernel function~\cite{rasmussenGaussianProcessesMachine2003}.
This is an advantage of GPs, as the most suitable features
for BLR are often challenging to determine. 
On the other hand, GPs scale poorly with the number of data samples \cite{lazaro-gredillaMarginalizedNeuralNetwork2010}.
For big data problems they are typically approximated with sparse GPs~\cite{liuWhenGaussianProcess2020},
and can be further improved with deep kernel learning~\cite{wilsonDeepKernelLearning2016, liuDeepLatentVariableKernel2022}.

Especially in recent years, neural networks (NNs) and deep learning have gained significant popularity for a vast variety of machine learning tasks \cite{Lecun2015}.
NNs have also been successfully applied 
for control applications~\cite{karg_Efficient_2020, dongFunctionalNonlinearModel2019},
often to infer the system model from data~\cite{sarangapaniSystemIdentificationUsing2006, fiedlerEconomicNonlinearPredictive2020, fiedlerModelPredictiveControl2022}.
A challenge with NNs is their tendency to overfit 
and their inability to express uncertainty~\cite{jospinHandsonBayesianNeural2022}.
Bayesian neural networks (BNNs),
in which the weights and predictions are probability distributions,
are a 
concept to tackle this shortcoming.
In practice, BNNs can be intractable to train and query
and are often approximated~\cite{jospinHandsonBayesianNeural2022, abdar_review_2021}.
Most approximate BNN approaches fall into one of two categories:
Markov chain Monte Carlo (MCMC) and variational inference (VI)~\cite{jospinHandsonBayesianNeural2022}.
MCMC methods do not require a classical training phase and instead
sample directly the posterior weight distribution 
of the Bayesian neural network~\cite{salakhutdinovBayesianProbabilisticMatrix2008}.
Unfortunately, MCMC scales poorly to large models which 
limits their applicability~\cite{jospinHandsonBayesianNeural2022}.
Variational inference is a popular alternative to MCMC 
and based on the idea of learning the parameters of a surrogate 
posterior distribution of the weights~\cite{jospinHandsonBayesianNeural2022, abdar_review_2021}.
For neural networks, variational inference is often implemented with the 
\emph{Bayes by Backprop} algorithm~\cite{blundell_Weight_2015}.

Unfortunately, even a BNN trained with VI requires 
sampling to approximate the predictive distribution~\cite{jospinHandsonBayesianNeural2022}.
This can be a significant 
disadvantage
in comparison to GPs and BLR, which yield analytical results.
A promising compromise 
between tractability and expressiveness 
are NNs with Bayesian last layer (BLL) \cite{lazaro-gredillaMarginalizedNeuralNetwork2010, watsonLatentDerivativeBayesian2021}. 
NNs with BLL
can be seen as a simplified BNN,
where only the  weights of the output layer follow a Gaussian distribution,
and 
the remaining 
layers contain deterministic weights. 
At the same time, they can be interpreted as a deep kernel learning
approach with linear kernel.
NNs with BLL are also strongly related to BLR 
in that they consider a nonlinear feature space which is mapped linearly onto the outputs. 
Similarly to GPs, BLR and other probabilistic models~\cite{rasmussenGaussianProcessesMachine2003,  bishopPatternRecognitionMachine2006}, 
NNs with BLL are trained by maximizing the marginal likelihood 
with respect to the parameters 
of the probabilistic model.
These parameters include prior and noise variance and, importantly, 
the weights of the deterministic layers.
While the log-marginal likelihood (LML) can be expressed analytically for NNs with BLL, 
it contains expressions such as the inverse of the precision matrix,
making it unsuitable for direct gradient-based optimization.
Previous works, especially for control applications, have therefore either 
assumed knowledge of all parameters \cite{harrisonControlAdaptationMetalearning2018, wabersichNonlinearLearningbasedModel2021, fiedlerModelPredictiveControl2022}, 
including prior and noise variance, or have maximized the LML 
after training a NN with fixed features \cite{mckinnonMetaLearningPaired2021, snoekScalableBayesianOptimization2015a}.
In previous works that did include the deterministic weights as parameters, 
maximizing the LML required sampling the surrogate posterior during training~\cite{watsonLatentDerivativeBayesian2021},
or using an approximate precision matrix~\cite{wilsonDeepKernelLearning2016} 
to enable gradient-based optimization. 

As a main contribution of this work, we propose an approach to maximize the exact LML of a NN with BLL that does not require sampling and is suitable for gradient-based optimization.
Most importantly, we avoid the matrix inverse in the LML
by reintroducing the weights of the last layer, which were marginalized, as optimization variables.
We show that our reformulation of the LML satisfies the conditions of optimality
for the same solution as the original formulation.
In this way, we provide a simpler training procedure, in comparison to training with variational inference,
and can outperform BLR with fixed features obtained from a NN.
Our second main contribution is an algorithm to improve the uncertainty quantification for extrapolation points.
To this end, we relate the computation of the BLL covariance to an intuitive metric for extrapolation, 
which is inspired by the definition of extrapolation in \cite{balestrieroLearningHighDimension2021}.
Based on this relationship, 
the proposed algorithm adjusts a scalar parameter to improve the log-predictive density on additional validation data.
Our proposed methods are derived for the multivariate-case,
estimating individual noise-variances for each output.
This is in contrast to GP and BLR applications, where the multivariate case 
is typically tackled by fitting an individual model to each output
\cite{mckinnonLearningProbabilisticModels2019, hewingCautiousModelPredictive2020, mckinnonMetaLearningPaired2021} 
or by assuming i.i.d. noise for all outputs~\cite{wabersichNonlinearLearningbasedModel2021}.
The advantage of a single model for the multivariate case is apparent 
for the application of system identification for optimization-based control,
as multiple models may significantly increase computation times.
%All of the presented algorithms and our results are available on online\footnote{\url{https://github.com/4flixt/2022_Paper_BLL_LML}}.

This work is structured as follows. 
In Section~\ref{sec:BLL_Introduction}, we introduce NNs with BLL.
The marginal likelihood, which is maximized during training,
is discussed
in Section~\ref{sec:Type-2-Likelihood}.
In Section~\ref{sec:Inter_extrapolation_BLL}, we discuss
interpolation and extrapolation for NNs with BLL
and present an algorithm to improve the predictive distribution in the extrapolation regime.
We discuss the special considerations required for the multivariate case in Section~\ref{sec:MultivariateCase}.
The Bayes by Backprop method
is introduced in Section~\ref{sec:BayesByBackprop}.
A simulation example to compare the proposed method
with BLR and Bayes by Backprop 
is presented in Section~\ref{ssec:Multivariate_toy_example}.
The paper is concluded in Section~\ref{sec:Conclusions}.

\section{Bayesian last layer}\label{sec:BLL_Introduction}\noindent 
We investigate a dataset $\gD=\lbrace\mX, \vt\rbrace$
consisting of $m$ data pairs of inputs $\vx\in\sR^{n_x}$ 
and targets $t\in\sR$ from which the matrices $\mX= \left[\vx_1,\dots,\vx_m\right]^\top\in\sR^{m\times n_x}$ and 
$\vt= \left[t_1,\dots,t_m\right]^\top\in\sR^{m\times 1}$ are formed.
We assume a scalar output for ease of notation and address the multivariate case in Section~\ref{sec:MultivariateCase}.
Regarding the notation, lower case bold symbols denote vectors,
upper case bold symbols denote matrices, and regular lower case symbols are scalars.
For the regression task, we introduce a feed-forward NN model
with $L$ hidden layers as
\begin{equation}
	y = NN(\vx; \sW_{L+1}) = g_{L+1} \circ h_{L+1} \circ \dots \circ g_1 \circ h_1(\vx),
	\label{eq:NN_formulation}
\end{equation}
where $\circ$ denotes function composition. 
At each layer, 
we have a linear mapping $h_l(\cdot)$ followed by a nonlinear activation function $g_l(\cdot)$,
that is:
\begin{align}
	\vh^\top_l & =  h_l(\va_{l-1}) = \left[\va_{l-1} ^\top,\ 1 \right]\mW_l & l\in\sI_{[1,L+1]},\\
	\va_l &= g_l(\vh_l) & l\in\sI_{[1,L+1]}.
\end{align}
The set $\sI_{[1,L+1]}$ denotes the set of integers from $1$ to $L+1$.
We have $\va_0 = \vx$, $\va_{L+1} = \vy$, 
and $\va_l\in\sR^{n_{a,l}}$ for all $l\in\sI_{[1,L]}$. 
The number of neurons in layer $l$ is denoted as $n_{a,l}$,
and we have the weights $\mW\in\sR^{n_{a,l-1}+1\times n_{a,l}}$, 
which include the bias term.
The set of weight matrices  with cardinality $L+1$ is denoted
$\sW_{L+1} = \{\mW_1,\dots,\mW_{L+1}\}$.
As a requirement for NNs with BLL, we state the following assumption.
\begin{assumption}\label{ass:NN_linear_activation_function}
	The NN  \eqref{eq:NN_formulation} has a linear activation function in the output layer, i.e. $g_{L+1}(h_{L+1})=h_{L+1}$.
\end{assumption}
With the linear mapping, the output of the last internal layer is of particular importance
and we introduce the notation:
\begin{align}\label{eq:Features_vs_augmented_Features}
	\tilde\vphi = \va_{L},\text{ and } \vphi = [\va_L^\top,\ 1]^\top,
\end{align}
where $\tilde\vphi\in\sR^{n_{\tilde\phi}}$ are referred to as \emph{linear features} and  $\vphi\in\sR^{n_{\phi}}$ are the corresponding \emph{affine features} with $n_{\phi}=n_{\tilde\phi}+1$.
With slight abuse of notation, we use $\tilde\vphi = \tilde\vphi(\vx;\sW_L)$ 
and  $\vphi = \vphi(\vx;\sW_L)$ as both, 
the values of the features, and the function, parameterized with $\sW_L$.
We require two additional assumptions to state Lemma~\ref{lem:BLL} which formally describes a NN with BLL.
\begin{assumption}\label{ass:NN_feature_space}
	The NN  \eqref{eq:NN_formulation} provides a  feature space $\vphi(\vx;\sW_L)\in\sR^{n_{\phi}}$
	from which the targets are obtained through a linear mapping, according to Assumption~\ref{ass:NN_linear_activation_function}:
	\begin{align}	\label{eq:NN_linear_feature_regression}%
		\vt= \vphi\left(\mX; \sW_L\right)^\top \vw + \rvepsilon = \vy + \rvepsilon,
	\end{align}
	where we introduce the set $\sW_L = \{\mW_1,\dots,\mW_L\}$ for all weights until layer $L$ and have  $\vw=\mW_{L+1}$ the weights of the output layer.
	The additive noise $\vepsilon\in\sR^{m}$ is zero-mean normally distributed, that is,
	$\rvepsilon\sim \gN(0,\mSigma_E)$.
\end{assumption}
\begin{assumption}\label{ass:NN_last_layer_weights_prior}
	We have a prior \emph{belief} for the weights of the output layer
	$\rvw \sim\gN(0,\mSigma_{\mW})$.
\end{assumption}
%This is exactly the setting of BLR \cite{bishopPatternRecognitionMachine2006},
We introduce 
the parameters 
of the probabilistic model~\eqref{eq:NN_linear_feature_regression}~as:
\begin{equation}\label{eq:Hyperparameters}
	\sTheta = \left\lbrace
	\sW_L , \mSigma_{\mE}, \mSigma_{\mW}
	\right\rbrace,
\end{equation}
and state the posterior distribution of the weights $\rvw$ 
of the last layer
using Bayes' law:
\begin{equation}\label{eq:Posterior_Bayesian_Linear_Regression}
	p(\rvw|\gD, \sTheta)=\frac{
		p(\gD| \rvw, \sTheta)p(\rvw|\sTheta)
	}
	{
		p(\gD| \sTheta)
	}.
\end{equation}
A neural network with Bayesian last layer 
has deterministic weights in all hidden layers
and distributed weights in the output layer.
We obtain
an analytical expression for
the distribution of the 
the predicted outputs 
as shown in the following lemma~\cite{watsonLatentDerivativeBayesian2021}.
\begin{lemma}\label{lem:BLL}
	Assumptions~\ref{ass:NN_linear_activation_function}-\ref{ass:NN_last_layer_weights_prior}
	hold.
	The predicted outputs are normally distributed with:
	\begin{subequations}
		\begin{align}
			p(\rvy|\gD ,\sTheta, \vx) &=
			\mathcal{N}(\vmu_{ y}^{\NN}(\vx), \mSigma_{ y}^{\NN}(\vx)),
			\label{eq:NN_BLL_distribution_a}\\
			\vmu_{ y}^{\NN}(\vx)&= \text{NN}(\vx; \sW_{L+1}),\\
			\mSigma_{ y}^{\NN}(\vx) &= \vphi^\top \mLambda_p^{-1} \vphi,
			\label{eq:NN_BLL_distribution_c}
		\end{align}
		\label{eq:NN_BLL_distribution}%
	\end{subequations}
	where $\mPhi = \vphi(\mX;\sW_L)$ is the feature matrix for the training data, 
	$\vphi = \vphi(\vx;\sW_L)$ is the feature matrix for the  test data,
	and with the precision matrix
	\begin{equation}\label{eq:Lambda_p_posterior}
		\mLambda_p = \mPhi^\top \mSigma_{\mE}^{-1} \mPhi + \mSigma_{\mW}^{-1}.
	\end{equation}
	%	 is the precision matrix.
	\begin{proof}
		The posterior $p(\rvw|\gD, \sTheta)$ in \eqref{eq:Posterior_Bayesian_Linear_Regression} yields a normal distribution in the weights:
		\begin{subequations}\label{eq:Posterior_Bayesian_Linear_Regression_weights}
			\begin{align}
				p(\rvw|\gD, \sTheta) &= \mathcal{N}(\bar\vw,\mLambda_p^{-1}),\\
				\text{with:}\quad
				\label{eq:w_bar_posterior}
				\bar\vw&=\mLambda_p^{-1}\mPhi^\top\mSigma_{\mE}^{-1}\vt,
			\end{align}%
		\end{subequations}
		as shown for the case of arbitrary features in \cite{bishopPatternRecognitionMachine2006}.
		Finally, the predicted output $\vy$ is a linear transformation of the random variable $\rvw$, yielding the distribution in \eqref{eq:NN_BLL_distribution}.
	\end{proof}
\end{lemma}
We can also obtain a posterior distribution for the targets,
for which we consider~\eqref{eq:NN_linear_feature_regression} and obtain
\begin{subequations}
	\begin{align}
		p(\rvt|\gD ,\sTheta, \vx) &=
		\mathcal{N}(\vmu_{ y}^{\NN}(\vx), \mSigma_{t}^{\NN}(\vx)),\\
		\mSigma_{t}^{\NN}(\vx) &= \mSigma_{y}^{\NN}(\vx) + \mSigma_{\mE},
	\end{align}
	\label{eq:NN_BLL_distribution_targets}%
\end{subequations}
where $\vmu_{ y}^{\NN}(\vx)$ and $\mSigma_{y}^{\NN}(\vx)$ stem from~\eqref{eq:NN_BLL_distribution}.
We will revisit~\eqref{eq:NN_BLL_distribution_targets} again for the definition of performance metrics.

\section{Marginal likelihood maximization}\label{sec:Type-2-Likelihood}\noindent
The posterior distribution of the weights
of the last layer can be obtained with~\eqref{eq:Posterior_Bayesian_Linear_Regression_weights}
for given values
of the parameters~$\sTheta$.
Determining suitable values of the parameters from prior knowledge
is often challenging.
Instead, the parameters can be inferred 
by maximizing the marginal likelihood in~\eqref{eq:Posterior_Bayesian_Linear_Regression},
which is also known as empirical Bayes or type-2 maximum likelihood~\cite{bernardoBayesianTheory1994a}.
In BLR, that is, for a fixed feature space, 
the log-marginal likelihood (LML) can be maximized as shown in~\cite{bishopPatternRecognitionMachine2006}.
Following this idea, the authors in \cite{snoekScalableBayesianOptimization2015a} propose an approach 
where a fixed feature space is obtained through classical NN regression.
After NN training, the LML is then maximized with these fixed features.
However, 
the LML is also influenced by the weights of the deterministic layers
and the resulting feature space. 
It is therefore reasonable to include the set of weights $\sW_L$ until the last layer as parameters
and train them by maximizing the LML.
Unfortunately, this poses significant challenges.

We proceed by introducing the LML and discuss 
the challenge of maximizing this expression in the next subsection.

\begin{lemma}[Log-marginal likelihood]\label{lem:LogMarginalLikelihood_formulation}
	If Assumptions~\ref{ass:NN_linear_activation_function}-\ref{ass:NN_last_layer_weights_prior} hold, 
	the negative LML of \eqref{eq:Posterior_Bayesian_Linear_Regression}, denoted as $J(\sTheta;\gD)=-\log(\gD| \sTheta))$, results in:
	\begin{equation}\label{eq:LogMarginalLikelihood}
		\small
		\begin{gathered}
			J(\sTheta;\gD)=
			\frac{m}{2}\log(2\pi)
			+\frac{1}{2}\log\det (\mSigma_\mW )
			+\frac{1}{2}\log\det (\mSigma_E)\\
			+\frac{1}{2}\log\det (\mLambda_p)
			+\frac{1}{2}\|\vt-\vy\|^2_{\mSigma_E^{-1}}
			+\frac{1}{2}\|\bar\vw\|^2_{\mSigma_\mW^{-1}}.
		\end{gathered}
	\end{equation}
	In this formulation, we have $\mLambda_p$ from  \eqref{eq:Lambda_p_posterior}, $\bar\vw$ from \eqref{eq:w_bar_posterior},
	$\vy=\mPhi\bar\vw$ and 
	$\sTheta$ from~\eqref{eq:Hyperparameters}.
	\begin{proof}
		The proof for fixed features is shown in \cite{bishopPatternRecognitionMachine2006}.
	\end{proof}
\end{lemma}
For practical applications, the formulation in~\eqref{eq:LogMarginalLikelihood} 
is further simplified by considering the following assumption, 
where $\diag(\mA,\mB)$ denotes a block-diagonal matrix with 
$\mA$ and $\mB$ on the diagonal.
\begin{assumption}\label{ass:Noise_prior_scalar}
	The additive noise introduced in Assumption~\ref{ass:NN_feature_space} is i.i.d. for all samples $m$, yielding $\mSigma_E=\sigma_e^2 \mI_m$.
	The weight prior introduced in Assumption~\ref{ass:NN_last_layer_weights_prior}
	is i.i.d. with a flat prior for the bias, that is,
	$\mSigma^{-1}_w=\sigma_w^{-2}\diag(\mI_{n_{\tilde\phi}},\ 0)$. 
\end{assumption}
The assumption of a flat prior for the bias term will have a negligible effect in practice
but is important for the theoretical results presented in Section~\ref{sec:Inter_extrapolation_BLL}.
For ease of notation,
we will reuse $J(\sTheta;\gD)$ and $\sTheta$ 
in following results.
\begin{res}\label{res:LogMarginal_Likelihood_SimplifiedFormulation}
	Applying Assumption~\ref{ass:Noise_prior_scalar} to Lemma~\ref{lem:LogMarginalLikelihood_formulation},
	we can write the negative LML in~\eqref{eq:LogMarginalLikelihood} as:
	\begin{equation}\label{eq:LogMarginalLikelihood_scalar_noise}
		\small
		\begin{gathered}
			J(\sTheta;\gD)=\frac{m}{2}\log (2\pi)
			+n_\phi\log(\sigma_w)+m\log(\sigma_e)\\
			+\frac{1}{2}\log\det (\mLambda_p)
			+\frac{1}{2\sigma_e^2}\|\vt-\vy\|^2_2
			+\frac{1}{2\sigma_w^2}\|\bar\vw\|^2_2.
		\end{gathered}
	\end{equation}
	We introduce $\tilde\mI_{n_\phi} = \diag(\mI_{n_{\tilde\phi}},\ 0)$
	and obtain simplified expressions for 
	\eqref{eq:Lambda_p_posterior} and \eqref{eq:w_bar_posterior}:
	\begin{subequations}
		\begin{align}
			\label{eq:Lambda_p_posterior_scalar}
			\mLambda_p &= \sigma_e^{-2}\mPhi^\top  \mPhi + \frac{1}{\sigma_w^2} \tilde\mI_{n_\phi},\\
			\label{eq:w_bar_posterior_scalar}
			\bar\vw&=\sigma_e^{-2}\mLambda_p^{-1}\mPhi^\top\vt.
		\end{align}
	\end{subequations}
	In this setting, we denote $\sTheta = \left\lbrace
	\sW_L , \sigma_e, \sigma_w
	\right\rbrace$.
\end{res}

\subsection{Augmented log-marginal likelihood maximization}\label{ssec:Augmented_LML}\noindent
To train a NN with BLL we seek to minimize the negative LML:
\begin{equation}\label{eq:LMLH_Optim_basic}
	\begin{aligned}
		\min_{\sTheta}\quad &J(\sTheta;\gD),\\
	\end{aligned}
\end{equation}
with  $J(\sTheta; \gD)$ and $\sTheta$ according to Result~\ref{res:LogMarginal_Likelihood_SimplifiedFormulation}.
This problem excludes the weights in the last layer of the NN as they have been marginalized. 
A result of this marginalization is the expression~\eqref{eq:w_bar_posterior_scalar} which computes these weights explicitly. 
Unfortunately, this creates a major challenge when iteratively solving problem \eqref{eq:LMLH_Optim_basic} for the optimal values $\sTheta^*$, as
the computational graph contains unfavorable expressions such as the inverse of $\mLambda_p$, which are both numerically challenging and computationally expensive. 
Furthermore, \eqref{eq:w_bar_posterior_scalar} requires leveraging the entire training dataset for the computation of the gradient $\nabla_{\sTheta}J(\sTheta;\gD)$.

The authors in \cite{watsonLatentDerivativeBayesian2021} circumvent these issues by variational inference, replacing the LML objective by the evidence lower bound objective (ELBO).
As the variational posterior they choose a Gaussian distribution, parameterized with mean and covariance.
This is the obvious choice in the BLL setting where the true posterior is Gaussian as shown in~\eqref{eq:NN_BLL_distribution} and,
consequently, the ELBO objective is equivalent to the LML \cite{watsonLatentDerivativeBayesian2021}. 
Variational inference comes at the cost, however, of introducing as optimization variables the parameters to describe the variational posterior.
Furthermore, the variational inference training loop requires sampling this variational posterior.

In this work, we present an alternative approach, 
which simultaneously avoids parameterizing the variational posterior 
and yields a computational graph with lower complexity than~\eqref{eq:LogMarginalLikelihood_scalar_noise}. 
The resulting formulation is suitable for fast gradient-based optimization.
To obtain this result we reformulate \eqref{eq:LMLH_Optim_basic} as
\begin{equation}\label{eq:LMLH_Optim_constrained}
	\begin{aligned}
		\min_{\sTheta, \bar\vw}&\quad J(\sTheta, \bar\vw; \gD)\\
		&\begin{aligned}
			&\text{s.t.}\quad  \bar\vw=\frac{1}{\sigma_e^2}\mLambda_p^{-1}\mPhi^\top\vt,
		\end{aligned}
	\end{aligned}
\end{equation}
with $\sTheta = \left\lbrace
\sW_L , \sigma_e, \sigma_w
\right\rbrace$.
Importantly, we introduce $\bar\vw$ as an optimization variable 
and add an equality constraint corresponding to \eqref{eq:w_bar_posterior_scalar}.
The optimal solution $\sTheta^*$ of \eqref{eq:LMLH_Optim_constrained} is thus identical to the optimal solution of \eqref{eq:LMLH_Optim_basic}.
As a main contribution of this work, we state the following theorem.
\begin{theorem}[Augmented log-marginal likelihood maximization]\label{theo:Aug_LMLH_Optimization}
	The optimal solution $(\sTheta^*, \bar\vw^*)$ of problem~\eqref{eq:LMLH_Optim_constrained} 
	is
	identical 
	to the optimal solution obtained from the unconstrained problem:
	\begin{equation}\label{eq:LMLH_Optim_unconstrained}
		\min_{\sTheta, \bar\vw}\quad J(\sTheta, \bar\vw; \gD),
	\end{equation}
	where, as the only difference to \eqref{eq:LMLH_Optim_basic},  $\bar\vw$ is now an optimization variable and, 
	in comparison to \eqref{eq:LMLH_Optim_constrained}, the equality constraint has been dropped.
	\begin{proof}
		The Lagrangian of problem~\eqref{eq:LMLH_Optim_constrained} can be written as:
		\begin{equation*}%\label{eq:Proof_Unconstrained_LMLH_1}
			\begin{gathered}
				\mathcal{L}(\sTheta, \bar\vw,\lambda; \gD)
				=J(\sTheta, \bar\vw; \gD)
				+\lambda^\top\left(\mLambda_p \bar\vw - \frac{1}{\sigma_e^2}\mPhi^\top\vt\right).
			\end{gathered}
		\end{equation*}
		We then state the first-order condition of optimality for the optimization variable $\bar\vw$:
		\begin{equation}\label{eq:Proof_Unconstrained_LMLH_2}
			\begin{gathered}
				\nabla_{\bar\vw}\mathcal{L}(\sTheta, \bar\vw,\lambda; \gD)
				=\nabla_{\bar\vw}J(\sTheta, \bar\vw; \gD)
				+\lambda^\top \mLambda_p  \overset{!}{=} 0.
			\end{gathered}
		\end{equation} 
		Considering~\eqref{eq:LogMarginalLikelihood_scalar_noise} and~\eqref{eq:NN_linear_feature_regression}, 
		that is, $\vy = \mPhi\bar\vw$, we obtain:
		\begin{align}
			\nabla_{\bar\vw}J&(\sTheta, \bar\vw; \gD)\nonumber\\
			&=\nabla_{\bar\vw}\left(
			\frac{1}{2\sigma_e^2}\|\vt-\mPhi\bar\vw\|^2_2
			+\frac{1}{2\sigma_w^2}\|\bar\vw\|^2_2
			\right)\\
			\label{eq:Proof_Unconstrained_LMLH_3}
			&= 
			\frac{2}{2\sigma_e^2}\left(
			\bar\vw^\top\mPhi^\top\mPhi-\vt^\top\mPhi\right)
			+\frac{2}{2\sigma_w^2}\bar\vw^\top\\
			\label{eq:Proof_Unconstrained_LMLH_4}
			&=
			\bar\vw^\top\left(\frac{1}{\sigma_e^2}\mPhi^\top  \mPhi + \frac{1}{\sigma_w^2} \tilde\mI_{n_\phi}\right)
			- \frac{1}{\sigma_e^2}\vt^\top\mPhi .
		\end{align}
		Using~\eqref{eq:Lambda_p_posterior_scalar}, we obtain:
		\begin{equation}\label{eq:Proof_Unconstrained_LMLH_5}
			\nabla_{\bar\vw}J(\sTheta, \bar\vw; \gD)
			=\bar\vw^\top\mLambda_p
			- \frac{1}{\sigma_e^2}\vt^\top\mPhi.
		\end{equation}
		We substitute~\eqref{eq:Proof_Unconstrained_LMLH_5} into~\eqref{eq:Proof_Unconstrained_LMLH_2} and have:
		\begin{align}
			\label{eq:Proof_Unconstrained_LMLH_6}
			\lambda^\top \mLambda_p &= -\bar\vw^\top\mLambda_p
			+\frac{1}{\sigma_e^2}\vt^\top\mPhi,\\
			\label{eq:Proof_Unconstrained_LMLH_7}
			\Leftrightarrow
			\lambda &= -\mLambda^{-1}_p\mLambda_p \bar\vw + \frac{1}{\sigma_e^2}\mLambda_p^{-1}\mPhi^\top\vt,\\
			\label{eq:Proof_Unconstrained_LMLH_8}
			\Leftrightarrow
			\lambda &= -\bar\vw + \bar\vw = 0, 
		\end{align}
		where in \eqref{eq:Proof_Unconstrained_LMLH_7} we have substituted \eqref{eq:w_bar_posterior_scalar}.
		From~\eqref{eq:Proof_Unconstrained_LMLH_8}, we obtain that $\lambda=0$ 
		and the Lagrangian of problem~\eqref{eq:LMLH_Optim_constrained} thus simplifies to:
		\begin{equation}
			\mathcal{L}(\sTheta, \bar\vw,\lambda; \gD) =J(\sTheta, \bar\vw; \gD).
		\end{equation}
		This is exactly the Lagrangian of problem~\eqref{eq:LMLH_Optim_unconstrained}
		and therefore problem~\eqref{eq:LMLH_Optim_constrained}
		and~\eqref{eq:LMLH_Optim_unconstrained} yield the same optimal solution. 
	\end{proof}
\end{theorem} 
Theorem~\ref{theo:Aug_LMLH_Optimization} enables us 
to maximize the LML~\eqref{eq:LogMarginalLikelihood_scalar_noise} 
without having to explicitly compute $\bar\vw$ according to Equation~\eqref{eq:w_bar_posterior_scalar}.
This means, in particular, that we are not required to compute the inverse of $\mLambda_p$ to express the LML.
Consequentially, we can use back-propagation and gradient-based optimization  
to maximize the LML, which significantly simplifies training NNs with BLL.

\subsection{Change of variables and scaling}\noindent
With Theorem~\ref{theo:Aug_LMLH_Optimization} we can state the LML 
as a function of $\bar\vw$ which is now included in the set of parameters
$\sTheta = \left\lbrace
\sW_{L}, \bar\vw,  a, b
\right\rbrace = \left\lbrace
\sW_{L+1} , a, b
\right\rbrace$.
Furthermore, we propose a change of variables and scale the objective function
to improve numerical stability.
In particular, we introduce:
\begin{equation}\label{eq:LMLH_alpha_definition}
	\alpha=\frac{\sigma_w^{2}}{\sigma_e^{2}},
\end{equation}
which can be interpreted as a signal-to-noise ratio 
and optimize over $\log(\alpha)$
and $ \log(\sigma_e)$
to ensure that $\sigma_e>0$ and $\sigma_w>0$ without constraining the problem. 
These changes are formalized in the following result.
\begin{res}\label{res:LML_Simplified_alpha_notation}
	Considering the definition of $\alpha$ in~\eqref{eq:LMLH_alpha_definition}, 
	we reformulate~\eqref{eq:Lambda_p_posterior_scalar}:
	\begin{equation}\label{eq:Lambda_p_vs_Lambda_p_bar}
		\begin{aligned}
			\mLambda_p &= \sigma_e^{-2}\mPhi^\top\mPhi+\sigma_w^{-2}\tilde\mI_{n_\phi}\\
			&= \sigma_e^{-2}\left(
			\mPhi^\top\mPhi+\alpha^{-1} \tilde\mI_{n_\phi}
			\right) = \sigma_{e}^{-2} \bar\mLambda_p,
		\end{aligned}
	\end{equation}
	where
	\begin{equation}\label{eq:Lambda_p_bar_alpha}
		\bar\mLambda_p = \mPhi^\top\mPhi+\alpha^{-1} \tilde\mI_{n_\phi}.
	\end{equation}
	The scaled negative LML from~\eqref{eq:LogMarginalLikelihood_scalar_noise} can then be written as:
	\begin{equation}\label{eq:LMLH_alpha_cost}
		\small
		\begin{gathered}
			J(\sTheta; \gD)=\frac{1}{2}\log (2\pi)
			+\frac{n_\phi}{2m}\log \alpha+\log \sigma_e\\
			+\frac{1}{2m}\left(
			\log\det(\bar\mLambda_p)
			+\sigma_e^{-2}\|\vt-\vy\|^2_2
			+\alpha^{-1}\sigma_e^{-2}
			\|\bar\vw\|^2_2\right),
		\end{gathered}
	\end{equation}
	with $\sTheta = \left\lbrace
	\sW_{L+1} , \alpha, \sigma_e
	\right\rbrace$. 
\end{res}
To train a NN with BLL and univariate output, we consider in the following
the LML and parameters in the form of Result~\ref{res:LML_Simplified_alpha_notation}.
Additionally, the newly introduced parameter $\alpha$ in~\eqref{eq:LMLH_alpha_definition}
will play an important role 
in the following discussion on interpolation and extrapolation
and ultimately helps to improve the extrapolative uncertainty.

\section{Improving the extrapolative uncertainty}\label{sec:Inter_extrapolation_BLL}\noindent
One of the main challenges of the BLL predictive distribution~\eqref{eq:NN_BLL_distribution} is that 
for arbitrary extrapolation points the required assumptions for Lemma~\ref{lem:BLL} will not hold.
In this section, we discuss the behavior of the predictive distribution in the interpolation and extrapolation regime 
and propose a method to improve the performance of NNs with BLL for extrapolation.

To formalize the notion of interpolation and extrapolation,
we follow the definition of interpolation described in~\cite{balestrieroLearningHighDimension2021},
%(thus implicitly defining extrapolation)
for which we also need to define the convex hull. 
\begin{defn}[Convex hull]\label{def:Convex_hull} 
	The convex hull of  a set of samples $\mX\in\sR^{m\times n_x}$ is defined as the set:
	\begin{equation*}
		\sC_{\mX} = \left\{
		\mX^\top \vnu | \vnu\in\sR^m,\  \sum \vnu=1,\vnu\geq0 
		\right\}.
	\end{equation*}
\end{defn}
\begin{defn}[Interpolation]\label{def:Interpolation}
	A sample $\vx$ is considered to be an interpolation point of a set of samples $\mX$, 
	given a feature space $\tilde\mPhi(\mX;\sW_L)$ which satisfies Assumption~\ref{ass:NN_linear_activation_function} 
	and~\ref{ass:NN_feature_space}, if:
	$
	%	\begin{equation*}
		\tilde\vphi(\vx)\in \sC_{\tilde\mPhi(\mX;\sW_L)}.
		%	\end{equation*}
	$
\end{defn}
Interpolation is an attribute of the input space but 
its definition
considers 
the learned feature space of the neural network.
Considering the feature space in Definition~\ref{def:Interpolation} may seem counter-intuitive but applies well to reality,
where nonlinear features for regression can show arbitrary behavior between data points, even for a univariate input. 
In this case, interpolation points in the input domain are rightly classified as extrapolation points by considering the feature domain. 

Classifying a point as interpolation or extrapolation is a binary decision. 
In reality this is a shortcoming,
as different degrees of extrapolation are possible. 
That is, a point \enquote{close} 
to the convex hull might still lead to a trustworthy prediction. 
The distance to a set, e.g. the convex hull,  is defined below.
\begin{defn}[Distance]\label{def:Distance_to_a_set}
	For a set $\sX\subset\sR^{n_x}$ and a point $\vx\in\sR^{n_x}$ we define the \emph{distance} $d(\vx,\sX) $ as:
	\begin{equation}
		d(\vx,\sX) = \inf_{a\in\sX} \|x-a\|_2^2.
	\end{equation}
\end{defn}

\subsection{Quantification of interpolation and extrapolation}\noindent\label{ssec:Affine_cost}
\begin{figure*}
	\centering
	\includegraphics[width=1\textwidth]{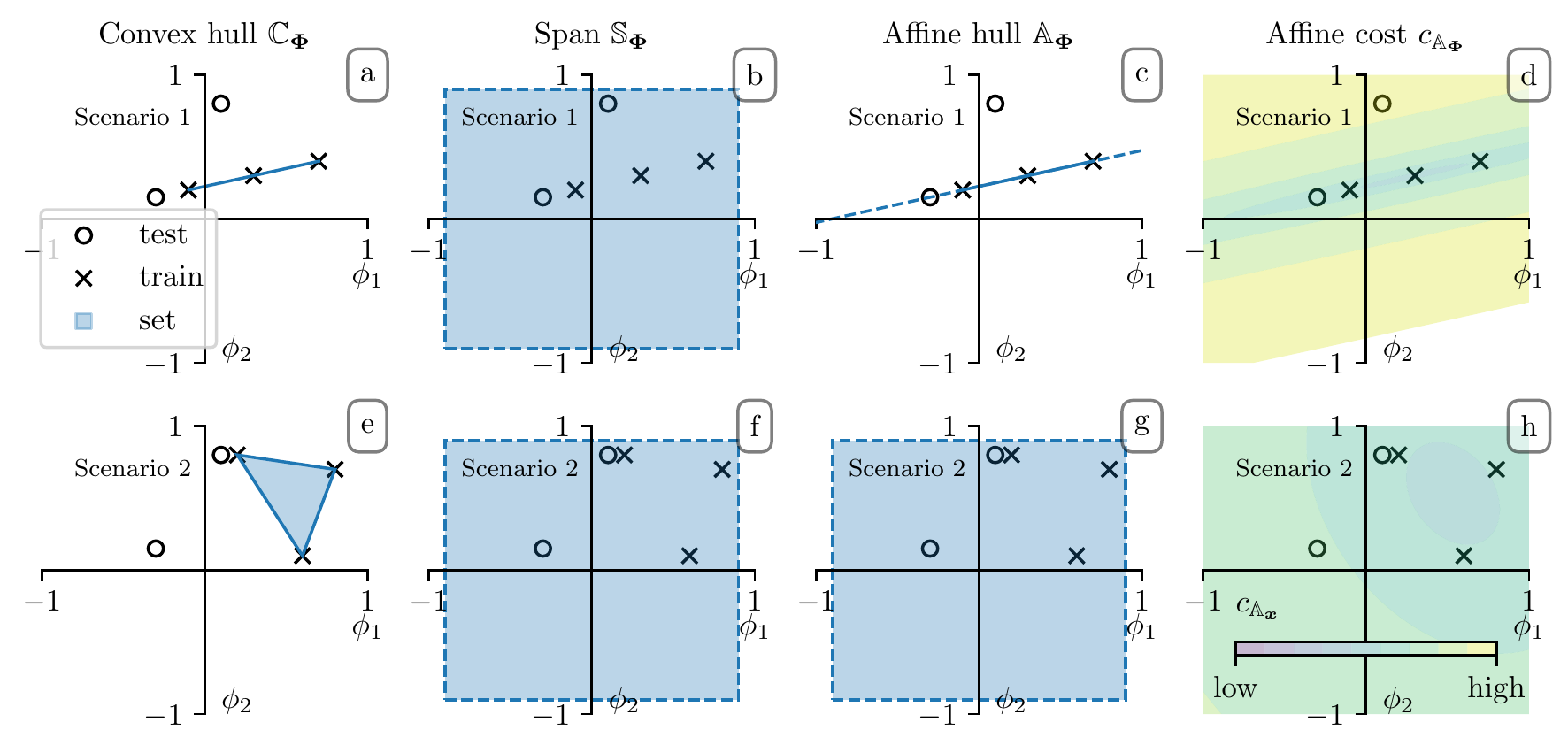}
	\caption{
		Comparison of convex hull (Definition~\ref{def:Convex_hull}), span (Definition~\ref{def:Span}),
		affine hull (Definition~\ref{def:Affine_hull}) and affine cost (Definition~\ref{def:Affine_cost})
		for two exemplary sets of features $\tilde\mPhi\in\sR^{m\times n_{\tilde\phi}}$, both with $m=3$ and $n_{\tilde\phi}=2$.
	}
	\label{fig:Convex_affine_span_hull_affine_cost}
\end{figure*}
In the following, we seek to define a metric to quantify the degree of extrapolation
for a nonlinear regression model with feature space.
Intuitively, such a metric could be based 
on the distance, according to Definition~\ref{def:Distance_to_a_set}, 
to the convex hull of the features.
Unfortunately, the distance to the convex hull results in an optimization problem that scales with the number of samples
and the feature dimension, which can be a limitation for practical applications.
Instead, we propose an approximate metric for which we introduce 
the \emph{affine cost}.
As related concepts of the affine cost, we define
the well known \emph{span} and \emph{affine hull}.
\begin{defn}[Span]\label{def:Span}
	The span of  a set of samples $\mX\in\sR^{m\times n_x}$ is defined as the set:
	\begin{equation*}
		\sS_{\mX} = \left\{
		\mX^\top \vnu | \vnu\in\sR^m\right\}.
	\end{equation*}
\end{defn}
\begin{defn}[Affine hull]\label{def:Affine_hull}
	The affine hull of  a set of samples $\mX\in\sR^{m\times n_x}$ is defined as the set:
	\begin{equation*}
		\sA_{\mX} = \left\{
		\mX^\top \vnu | \vnu\in\sR^m,\  \sum \vnu=1
		\right\}.
	\end{equation*}
\end{defn}
\begin{defn}[Affine cost]\label{def:Affine_cost}
	The affine cost of a a test point $\vx\in\sR^{n_x}$ and given the data $\mX\in\sR^{m\times n_x}$
	is defined as the optimal cost:
	\begin{subequations}\label{eq:Affine_cost}
		\begin{align}
			c_{\sA_\mX}(\vx) =  \minimize{\vnu, \ve}& \quad \|\vnu\|_2^2+ \gamma\|\ve\|_2^2\\
			\mathrm{subject\ to:}&\quad \mX^\top\vnu + \ve = \vx,\\
			&\sum \vnu = 1,
		\end{align}
	\end{subequations}
	where the spanning coefficient $\vnu\in\sR^m$ and the residual variable $\ve\in\sR^{n_x}$
	are optimization variables and $\gamma\in\sR$ is a weighting factor for the residuals.
\end{defn}
The affine cost naturally complements the definition of the affine hull
by computing the norm of the respective spanning coefficients $\vnu$ from Definition~\ref{def:Affine_hull}.
Importantly, test values $\vx\notin \sA_{\mX}$ also have a value assigned,
for which the $\gamma$-weighted norm of the residuals $\ve$ is considered.

The relationship of convex hull, span and affine hull as well as 
the affine cost can be inspected in Figure~\ref{fig:Convex_affine_span_hull_affine_cost}.
In this figure, two exemplary sets of features $\tilde\mPhi\in\sR^{m\times n_{\tilde\phi}}$, both with $m=3$ and $n_{\tilde\phi}=2$
are presented and we compare the relationship of test to training samples. 

By comparing, for example Figure~\ref{fig:Convex_affine_span_hull_affine_cost} a) and d), 
we can see that the affine cost
has a strong relationship with  the convex hull on which we have based the notion of interpolation in Definition~\ref{def:Interpolation}.
In particular, we consider the level set 
\begin{equation*}
	\sL_{\sA_\mPhi} =\{ \tilde\vphi\in\sR^{n_{\tilde\phi}} | c_{\sA_{\tilde\mPhi}}(\tilde\vphi)\leq l  \},
\end{equation*}
obtained with the affine cost and suitable level $l$ as a soft approximation of the convex hull. 
Such level sets can be seen in Figure~\ref{fig:Convex_affine_span_hull_affine_cost} d) and h) as the contour lines of $c_{\sA_{\tilde\mPhi}}$.
It holds that for a test point $\tilde\vphi$,
the affine cost grows with the distance to the level set $\sL_{\sA_\mPhi}$.  
In this sense, we consider the distance $d(\tilde\vphi, \sL_{\sA_\mPhi})$ 
and, more directly, 
the affine cost $c_{\sA_\mPhi}(\tilde\vphi)$ itself 
as the desired metric for the degree of extrapolation.

For the behavior of this metric we distinguish two important cases.
In the first case, small values of the affine cost are achieved 
for test points that are within the affine hull, i.e. $\tilde\vphi\in\sA_{\tilde\mPhi}$ , and for small distances to the convex hull.
According to Definition~\ref{def:Interpolation}, these include all interpolation points
and what we consider mild extrapolation.
Both test points in Figure~\ref{fig:Convex_affine_span_hull_affine_cost}~h) are examples for this case.

In the second case, a test point is not within the affine hull $\tilde\vphi\notin\sA_{\tilde\mPhi}$.
The affine cost is now influenced primarily through the parameter $\gamma$.
This can be seen by inspection of \eqref{eq:Affine_cost}, 
where the residuals $\ve$ must be used if $\tilde\vphi\notin\sA_{\tilde\mPhi}$.
The cost may then be dominated by the term $\|\ve\|_2^2$
which is weighted with $\gamma$.
We argue that in the second case the test point can be considered an extrapolation point 
and $\gamma$ can be interpreted as a penalty for extrapolation.
This case can be observed in Figure~\ref{fig:Convex_affine_span_hull_affine_cost}~d) for the test point with higher affine cost.

\subsection{Relationship of affine cost and covariance}\noindent
\begin{figure*}
	\centering
	\includegraphics[width=1\textwidth]{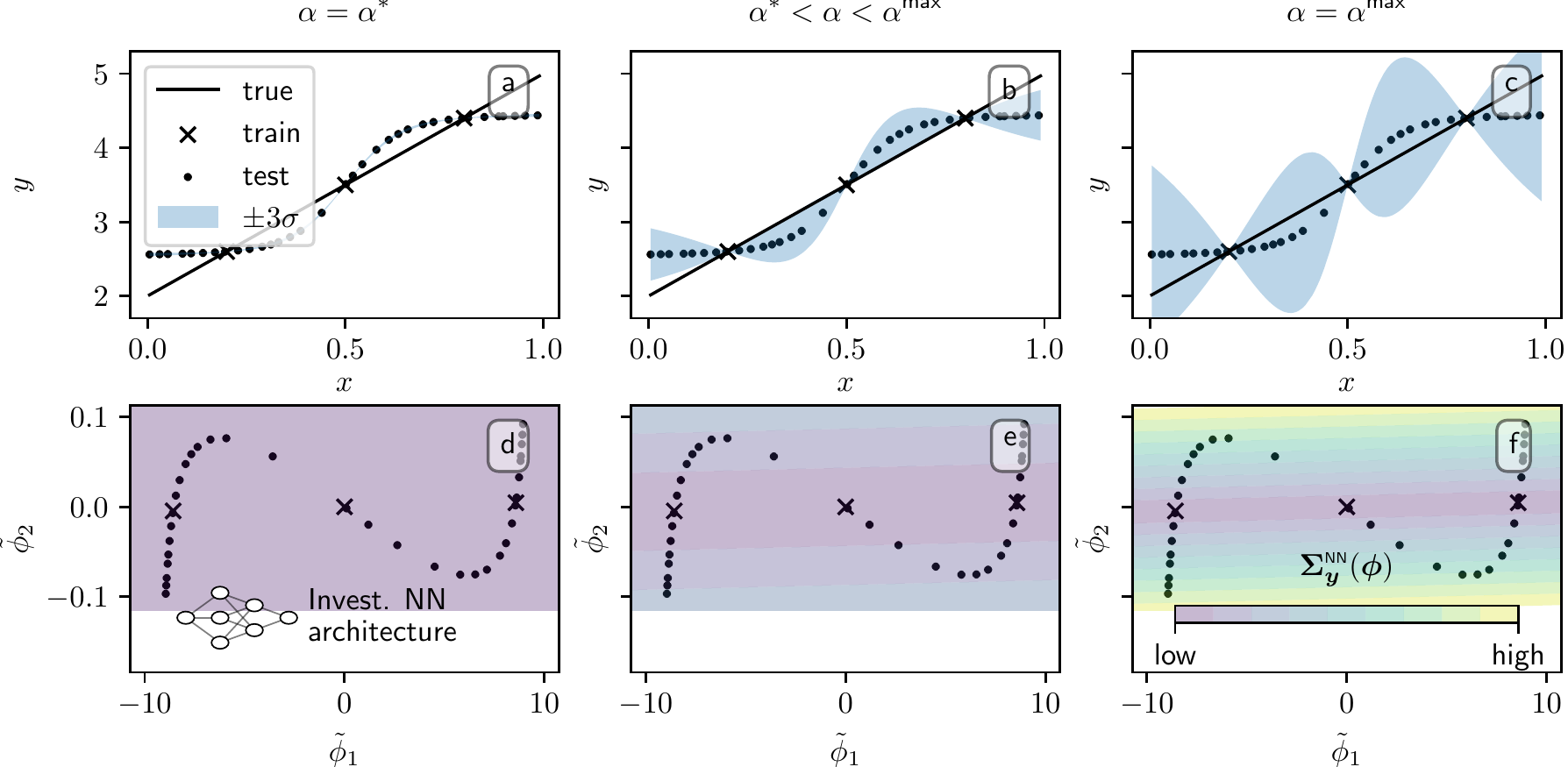}
	\caption{
		NN  with BLL: Predicted mean and standard deviation~\eqref{eq:NN_BLL_distribution} and feature space with $n_{\tilde{\phi}}=2$ for $m=3$ training samples.
		The effect of parameter $\alpha$ on the extrapolation uncertainty is shown
		by comparing the optimal $\alpha^*$ (maximization of LML~\eqref{eq:LMLH_alpha_cost}) with suggested improved $\alpha^{\max}$.
	}
	\label{fig:Funda_08_Pred_and_Featurespace_alpha_comparison}
\end{figure*}\noindent
As another main contribution of this work, we introduce Theorem~\ref{theo:BLL_Affine_cost}
to establish the relationship of the BLL covariance and the previously presented affine cost.

\begin{theorem}[BLL affine cost]\label{theo:BLL_Affine_cost}
	If Assumption~\ref{ass:Noise_prior_scalar} holds, 
	and with $\gamma=\alpha$,
	the affine cost $c_{\sA_{\tilde\mPhi}}(\tilde\vphi(\vx))$, according to Definition~\ref{def:Affine_cost}, 
	is equivalent to the scaled BLL covariance \eqref{eq:NN_BLL_distribution_c}:
	\begin{equation}\label{eq:BLL_Affine_cost}
		\begin{aligned}
			c_{\sA_{\tilde\mPhi}}(\tilde\vphi(\vx)) &= \sigma_e^{-2} \mSigma_{\hat y}^{\NN}(\vx),
		\end{aligned}
	\end{equation}
	where $\vphi$ and $\tilde\vphi$ 
	describe the features obtained at the last internal layer of the neural network as defined in~\eqref{eq:Features_vs_augmented_Features}.
	\begin{proof}
		We reformulate the equality constraints of $c_{\sA_{\tilde\mPhi}}(\tilde\vphi)$ shown in \eqref{eq:Affine_cost}:
		\begin{equation}\label{eq:BLL_Affine_cost_derivation_01}
			\begin{bmatrix}
				\tilde\mPhi^\top & \mI_{n_{\tilde\phi}}\\
				\mOne_{1,n_m} & \vzero_{1,n_{\tilde\phi}}
			\end{bmatrix}
			\begin{bmatrix}
				\vnu\\
				\ve
			\end{bmatrix}=
			\begin{bmatrix}
				\tilde\vphi\\
				1
			\end{bmatrix},
		\end{equation}
		where $\vzero$ and $\vone$ denote matrices filled with zeros or ones 
		and their respective dimensions are given in the subscript.
		Considering also~\eqref{eq:Features_vs_augmented_Features}, we then introduce:
		\begin{equation*}
			\mPhi^\top = \begin{bmatrix}
				\tilde\mPhi^\top\\
				\mOne_{1,n_m}
			\end{bmatrix},\
			\mM = \begin{bmatrix}
				\mI_{n_{\tilde\phi}}\\
				\vzero_{1,n_{\tilde\phi}}
			\end{bmatrix},\
			\vp = \begin{bmatrix}
				\vnu\\
				\ve
			\end{bmatrix}\text{ and }
			\vphi = \begin{bmatrix}
				\tilde\vphi\\
				1
			\end{bmatrix},
		\end{equation*}
		which allows to reformulate
		Equation~\eqref{eq:BLL_Affine_cost_derivation_01} as:
		\begin{align}
			\label{eq:BLL_Affine_cost_derivation_02}
			\begin{bmatrix}
				\mPhi^\top & \mM\\
			\end{bmatrix}
			\vp&=
			\vphi .
		\end{align}
		By further introducing 
		$ \mD =\begin{bmatrix}
			\mPhi^\top & \mM\\
		\end{bmatrix}$ 
		and
		$\mW = \diag(\mI_{m},\ \gamma \mI_{n_{\tilde\phi}})$,
		Problem~\eqref{eq:Affine_cost} can be stated as:
		\begin{equation}\label{eq:BLL_Affine_cost_derivation_04}
			\begin{aligned}
				c_{\sA_\mX}(\tilde\vphi) =  \minimize{\vp}& \quad \|\vp\|^2_\mW \\
				\mathrm{subject\ to:}&\quad \mD\vp = \vphi.
			\end{aligned}
		\end{equation}
		We have  a weighted least-squares problem in~\eqref{eq:BLL_Affine_cost_derivation_04}
		for which the solution can be obtained as:
		\begin{equation}\label{eq:BLL_Affine_cost_derivation_05}
			\vp^* = \vphi^\top(\mD\mW^{-1}\mD^\top)^{-1}D\mW^{-1}.
		\end{equation}
		Substituting $\vp^*$ from~\eqref{eq:BLL_Affine_cost_derivation_05} 
		into the cost function of~\eqref{eq:BLL_Affine_cost_derivation_04}
		yields the affine cost:
		\begin{equation}\label{eq:BLL_Affine_cost_derivation_06}
			\begin{aligned}
				c_{\sA_\mX}(\tilde\vphi) &= \vphi^\top(\mD\mW^{-1}\mD^\top)^{-1}\vphi,
			\end{aligned}
		\end{equation}
		where
		\begin{equation}\label{eq:BLL_Affine_cost_derivation_07}
			\begin{aligned}
				\mD\mW^{-1}\mD^\top &= 
				\begin{bmatrix}
					\mPhi^\top & \mM\\
				\end{bmatrix}
				\begin{bmatrix}
					\mI_m & \\
					&\gamma^{-1}\mI_{n_{\tilde\phi}}\\
				\end{bmatrix}
				\begin{bmatrix}
					\mPhi \\ \mM^\top\\
				\end{bmatrix}\\
				&= \mPhi^\top\mPhi +\gamma^{-1}\tilde\mI_{n_\phi}.
			\end{aligned}
		\end{equation}
		Considering that $\gamma = \alpha$, as stated in the theorem, 
		and~\eqref{eq:Lambda_p_vs_Lambda_p_bar}-\eqref{eq:Lambda_p_bar_alpha},
		we have:
		\begin{align}\label{eq:BLL_Affine_cost_derivation_08}
			\mD\mW^{-1}\mD^\top &= \mPhi^\top\mPhi +\alpha^{-1}\tilde\mI_{n_\phi}
			= \bar\mLambda_p = \sigma_e^2 \mLambda_p,
		\end{align}
		where $\tilde\mI_{n_\phi} = \diag(\mI_{n_{\tilde\phi}},\ 0)$.
		Substituting~\eqref{eq:BLL_Affine_cost_derivation_08} in~\eqref{eq:BLL_Affine_cost_derivation_06} yields:
		\begin{equation}
			c_{\sA_\mX}(\tilde\vphi(\vx)) = \sigma_e^{-2}\vphi^\top\mLambda_p^{-1}\vphi=\sigma_e^{-2} \mSigma_{\hat y}^{\NN}(\vx),
		\end{equation}
		which concludes the proof.
	\end{proof}
\end{theorem}
Theorem~\ref{theo:BLL_Affine_cost} establishes the theoretical relationship between affine cost and BLL covariance.
In the next subsection, we discuss how this relationship 
can be used to improve the extrapolative uncertainty of NNs with BLL.

\subsection{Improving the uncertainty quantification with  the affine cost interpretation}\noindent
The first practical implication of Theorem~\ref{theo:BLL_Affine_cost} is that a NN with BLL
should indeed be trained by maximizing the marginal likelihood as discussed in Section~\ref{sec:Type-2-Likelihood}.
Using the features of a trained NN and then applying BLR,
as previously shown in~\cite{mckinnonMetaLearningPaired2021, snoekScalableBayesianOptimization2015a} 
might lead to suboptimal performance.
The reason for this is the $\log\det$-regularization of the precision matrix $\mLambda_p$
which arises only in the marginal likelihood cost formulation. 
This regularization encourages a low rank of the feature matrix $\tilde\mPhi$, resulting in a proper subspace for the affine hull, that is,
$\sA_{\tilde\mPhi}\subset\sR^{n_{\tilde\phi}}$.
Only in this setting can we potentially obtain test points with $\tilde\vphi(\vx)\notin\sA_{\tilde\mPhi}$, 
clearly indicating extrapolation through high values of the affine cost.
The effect can be seen in Figure~\ref{fig:Convex_affine_span_hull_affine_cost} 
by comparing subplot d) and h).  
The application of $\log\det$-regularization to obtain matrices with low rank
is well known~\cite{fazelLogdetHeuristicMatrix2003} and also applied in other fields such as compressed sensing~\cite{dongCompressiveSensingNonlocal2014}.

The second important implication of Theorem~\ref{theo:BLL_Affine_cost} is the interpretation
of the parameter $\alpha$ (or $\gamma$ respectively) in the context of the affine cost from Definition~\ref{def:Affine_cost}.
The parameter directly controls the affine cost and thus the variance for test points $\tilde\vphi(\vx)\notin\sA_{\tilde\mPhi}$.
This causes a dilemma: Naturally, we have the situation that all training samples $\tilde\mPhi$ are within the affine hull of themselves. 
Therefore, extrapolation in the sense of $\tilde\vphi(\vx)\notin\sA_{\tilde\mPhi}$ does not occur during training 
and $\alpha^*$, which maximizes the LML, might not yield desirable results.

To illustrate the issue we present a simple regression problem with $n_x=n_y=1$ and $m=3$ samples. 
We investigate a NN with $n_{\tilde{\phi}}=2$  which allows for a graphical representation of the feature space.
The NN is trained by maximizing the LML~\eqref{eq:LMLH_alpha_cost}, yielding the optimal parameters $\sTheta^*$,
which includes $\alpha^*$.
The predicted mean and standard deviation for the trained model using $\alpha^*$ can be seen
in Figure~\ref{fig:Funda_08_Pred_and_Featurespace_alpha_comparison}~a). 
The predicted mean of the NN, in light of the sparse training data, is suitable.
However, the variance shows that the prediction is overconfident.
We show in Figure~\ref{fig:Funda_08_Pred_and_Featurespace_alpha_comparison}~b)
and~c) that this overconfidence can be tackled simply by increasing $\alpha$ 
relative to the optimal value $\alpha^*$.

The reason for this effect of $\alpha$ on the extrapolation uncertainty
can be seen by inspecting Figure~\ref{fig:Funda_08_Pred_and_Featurespace_alpha_comparison}~d)-f), 
where the features (recall $n_{\tilde{\phi}}=2$) for the test ($\tilde\vphi$) and the training points ($\tilde\mPhi$) are displayed.

As desired, we have that $\sA_{\tilde\mPhi}\subset\sR^{n_{\tilde\phi}}$, 
that is, the training features lie in a subspace of $\sR^2$,
which can be seen in Figure~\ref{fig:Funda_08_Pred_and_Featurespace_alpha_comparison}
where the training samples in the feature space could be connected by a straight line.
This effect can be attributed to the $\log\det$-regularization in the LML.
Extrapolation thus occurs for $\tilde\vphi\notin\sA_{\tilde\mPhi}$ 
and for these points the extrapolative uncertainty grows with increasing $\alpha$.
Importantly, by considering Definition~\ref{def:Interpolation}, 
we also have extrapolation for test points that are within the 
convex hull of the input space, i.e. $\vx\in\sC_{\mX}$.
In this example, the only true interpolation points are the training samples
for which
increasing $\alpha$ has no significant effect on the predicted variance.

\begin{figure}
	\centering
	\includegraphics[width=1\columnwidth]{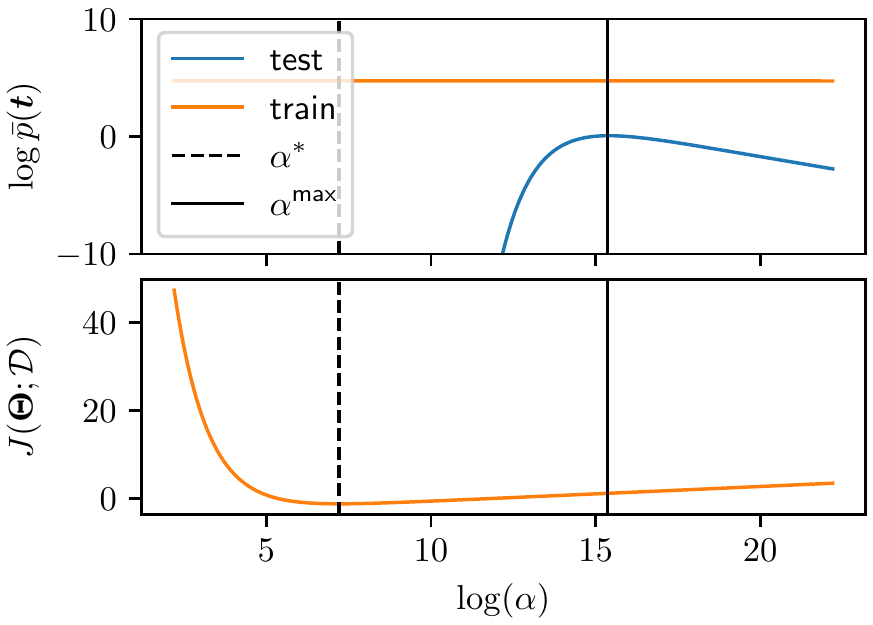}
	\caption{Effect of $\log(\alpha)$ on the LML for a trained NN and 
		the mean log-predictive density~\eqref{eq:Mean_predictive_probability}. 
		The same regression problem and NN as in  Figure~\ref{fig:Funda_08_Pred_and_Featurespace_alpha_comparison} is considered.
	} 
	\label{fig:Funda_08_logprob_lmlh_depence_alpha}
\end{figure}

In Figure~\ref{fig:Funda_08_logprob_lmlh_depence_alpha}, we further investigate the effect of increasing $\alpha$ for the same regression problem and 
NN architecture as displayed in Figure~\ref{fig:Funda_08_Pred_and_Featurespace_alpha_comparison}.
To quantify the quality of the predictive distribution, 
we use the log-predictive density (LPD) \cite{kerstingMostLikelyHeteroscedastic2007}:
\begin{equation}\label{eq:Mean_predictive_probability}
	\begin{gathered}
		\small
		\log \bar p(\vt^\test)=
		\frac{1}{m_\test}\sum_{i=1}^{m_\test}\log p(\rt=t^\test_i|\gD,\sTheta,\vx^\test),
	\end{gathered}
\end{equation}
which evaluates the logarithm of the posterior distribution of the targets~\eqref{eq:NN_BLL_distribution_targets}
for all test values and computes the average thereof.

In Figure~\ref{fig:Funda_08_logprob_lmlh_depence_alpha},
we display log-predictive density~\eqref{eq:Mean_predictive_probability}
and the negative LML~\eqref{eq:LMLH_alpha_cost}
for the optimal parameters $\sTheta^*$  and as a function of $\alpha$.
As expected, $\alpha^*$ minimizes the negative LML.
We see, however, that the optimal value is almost at a plateau and 
that increasing $\alpha$ has only a minor effect on the LML.
We also see in the top diagram of Figure~\ref{fig:Funda_08_logprob_lmlh_depence_alpha}
that the log-predictive density for the training points remains
independent of $\alpha$. 
On the other hand, we see that the log-predictive density for test points can be increased significantly 
by setting $\alpha=\alpha^{\text{max}}$.
Further increasing $\alpha$ has a negative effect 
as the predicted posterior distribution becomes increasingly flat. 

As a consequence of this investigation we propose Algorithm~\ref{alg:BLL_and_alpha} 
to obtain a NN with BLL, trained with LML maximization,
and enhanced extrapolation uncertainty.
\begin{algorithm}
	\caption{LML optimization with adapted extrapolation penalty $\alpha$ for enhanced BLL.}\label{alg:BLL_and_alpha}
	\begin{algorithmic}
		\Require $\gD^\train$, $\gD^\val$
		\Require NN structure ($L$, $n_{a_l}, g_l(\cdot)\ \forall l \in\sI_{[1,L]}$)
		\State $\sTheta^* \gets \arg\min_{\sTheta} J(\sTheta;\gD^\train)$ \Comment{solve \eqref{eq:LMLH_alpha_cost}}
		\State $\alpha^{\max} \gets  \arg\max_\alpha \log \bar p(\vt^\val)$ \Comment{solve \eqref{eq:Mean_predictive_probability}}
	\end{algorithmic}
\end{algorithm}

The optimal value for the scalar
parameter
$\alpha$ in Algorithm~\ref{alg:BLL_and_alpha} 
can be 
obtained
with a simple bijection approach. 

\section{The multivariate case}\label{sec:MultivariateCase}\noindent
In general, we seek to investigate multivariate problems where, in contrast to Section~\ref{sec:BLL_Introduction}, 
we now have that $\vt \in\sR^{n_y}$. 
As before, we consider a dataset $\gD=\lbrace\mX,\mT\rbrace$ consisting of $m$ samples and introduce 
$\mT=\left[\vt_1,\dots,\vt_m\right]^\top\in\sR^{m\times n_y}$.

To obtain the setting in which Lemma~\ref{lem:BLL} holds, 
we augment~\eqref{eq:NN_linear_feature_regression}
for the multivariate case:
\begin{equation}\label{eq:NN_linear_feature_regression_multivariate}
	\mT = \mPhi\mW+\mE,
\end{equation}
where $\mE\in\sR^{m\times n_y}$ is the matrix of residuals of the regression model.
We use the $\vec(\cdot)$ operation as defined in \cite[Defn. 11.5]{seber_matrix_2008} 
and vectorize \eqref{eq:NN_linear_feature_regression_multivariate}:
\begin{align}\label{eq:NN_linear_feature_regression_multivariate_vectorized_01}
	\vec(\mT) = \vec(\mPhi\mW) + \vec(\mE).
\end{align}
Introducing $\otimes$ as the Kronecker product,
this equation can be reformulated as \cite[Prop. 11.16(b)]{seber_matrix_2008}:
\begin{align}\label{eq:NN_linear_feature_regression_multivariate_vectorized_02}
	\vec(\mT) = (\mI_{n_y}\otimes\mPhi)\vec(\mW) + \vec(\mE),
\end{align}
and expressed as: 
\begin{equation}\label{eq:NN_linear_feature_regression_multivariate_vectorized_final}
	\hat\vt = \hat\mPhi\hat\vw+\hat\vepsilon.
\end{equation}
In the form of \eqref{eq:NN_linear_feature_regression_multivariate_vectorized_final}, Lemma~\ref{lem:BLL} directly applies to the multivariate case
where the prior and noise covariances from Assumption~\ref{ass:NN_feature_space} and~\ref{ass:NN_last_layer_weights_prior} now refer to:
\begin{align*}
	\hat\vepsilon&\sim \gN(0,\hat\mSigma_E),\\
	\hat\vw &\sim\gN(0,\hat\mSigma_{\mW}),
\end{align*}
and for which we need to consider the multivariate feature matrix $\hat\mPhi$.

\subsection{Simplified training}\noindent
The multivariate settings adds significant complexity 
to the marginal likelihood maximization introduced in Section~\ref{sec:Type-2-Likelihood}.
To reduce the computational complexity, we present Result~\ref{res:LMLH_multivariate_simplified} 
and two required assumptions in the following.
\begin{assumption}\label{ass:Noise_prior_scalar_multivariate}
	Noise and prior, as introduced in Assumption~\ref{ass:NN_feature_space} and~\ref{ass:NN_last_layer_weights_prior}, 
	are uncorrelated for all $n_y$ outputs and  identical for all $m$  samples and all $n_\phi$ features, respectively.
	We denote $\vsigma_e = \left[\sigma_{e,1},\ \dots,\ \sigma_{e,n_y}\right]$ and $\vsigma_w = \left[
	\sigma_{w,1},\ \dots,\ \sigma_{w,n_y}
	\right]$ and obtain:
	\begin{align}
		\label{eq:hatSigma_E_multivariate}
		\hat\mSigma_E &= \diag(\vsigma_e)\otimes \mI_m,\\
		\label{eq:hatSigma_W_multivariate}
		\hat\mSigma_W &= \diag(\vsigma_w)\otimes \mI_{n_\phi}.
	\end{align}
\end{assumption}
We introduce $\valpha=\frac{\vsigma_w^{2}}{\vsigma_e^{2}}$,
similarly to \eqref{eq:LMLH_alpha_definition}, and assume the following.
\begin{assumption}\label{ass:Alpha_constant}
	For all predicted outputs $y_i$, we have the same parameter $\alpha$, i.e.:
	\begin{equation}
		\alpha_1 = \dots = \alpha_{n_y}\overset{!}{=}\alpha.
	\end{equation}
\end{assumption}
%TODO: Fix here a and b.
\begin{res}\label{res:LMLH_multivariate_simplified}
	Let Assumption~\ref{ass:Noise_prior_scalar_multivariate} and~\ref{ass:Alpha_constant} hold.
	The scaled negative LML in \eqref{eq:LogMarginalLikelihood} 
	for the vectorized multivariate case~\eqref{eq:NN_linear_feature_regression_multivariate_vectorized_final} is:
	\begin{equation}\label{eq:LMLH_multivariate_simplified}
		\begin{aligned}
			\small
			J(\sTheta; \gD)&=
			\frac{n_y}{2m}\left(
			m\log (2\pi)
			+n_\phi \log(\alpha)+
			\log\det(\bar\mLambda_p)\right)\\
			&+
			\sum_{i=1}^{n_y}\left(
			\log(\sigma_{e,i})+
			\frac{1}{2m}\sigma_{e,i}^{-2}\|\vt_i-\vy_i\|^2_2
			\right)\\
			&+\sum_{i=1}^{n_y}\left(\frac{1}{2m}\alpha^{-1}\sigma_{e,i}^{-2}
			\|\bar\vw_i\|^2_2\right),
		\end{aligned}
	\end{equation}
	with  $\sTheta = \left\lbrace
	\sW_{L+1} , \alpha, \sigma_{e,1}, \dots, \sigma_{e,n_y}
	\right\rbrace$ and $\bar\mLambda_p$ according to \eqref{eq:Lambda_p_bar_alpha}.
	The full precision matrix $\hat\mLambda_p$ for the multivariate case can be obtained as:
	\begin{equation}
		\hat\mLambda_p = \diag(\vsigma_e^{-2})\otimes \bar\mLambda_p.
	\end{equation}
	\begin{proof}
		The result follows directly from the properties of the Kronecker product \cite{seber_matrix_2008}
		applied to the log-marginal likelihood in~\eqref{eq:LogMarginalLikelihood}
		with~\eqref{eq:NN_linear_feature_regression_multivariate_vectorized_final},
		as well as Assumption~\ref{ass:Noise_prior_scalar_multivariate}
		and~\ref{ass:Alpha_constant}.
	\end{proof}
\end{res}
Result~\ref{res:LMLH_multivariate_simplified} shows that the LML can 
be easily expressed for the multivariate case, allowing for fast and efficient NN training.
This is largely due to Assumption~\ref{ass:Noise_prior_scalar_multivariate} and~\ref{ass:Alpha_constant}.
While Assumption~\ref{ass:Noise_prior_scalar_multivariate} is a natural extension of
Assumption~\ref{ass:Noise_prior_scalar} for the multivariate case,
Assumption~\ref{ass:Alpha_constant} might be questioned.
We argue again with our  interpretation of $\alpha$ as an extrapolation penalty weight, 
as discussed in  Section~\ref{sec:Inter_extrapolation_BLL}.
Importantly, extrapolation, as defined in Definition~\ref{def:Interpolation}, 
is a property of the feature space and occurs regardless of the number of outputs.

Apart from simplifying the LML in~\eqref{eq:LMLH_multivariate_simplified},
Assumption~\ref{ass:Noise_prior_scalar_multivariate} and~\ref{ass:Alpha_constant} 
also yield a simplified computation of the predictive distribution.
The outputs are uncorrelated due to Assumption~\ref{ass:Noise_prior_scalar_multivariate} 
and we can obtain independent covariance matrices:
\begin{equation}
	\mSigma_{\hat y_i}^{\NN}(\vx) = \vphi^\top \mLambda_{p,i}^{-1} \vphi,\quad \text{with: }
	\mLambda_{p,i} = \sigma_{e,i}^{-2} \bar\mLambda_p,
\end{equation}
where we have the same $\bar\mLambda_p$ for all outputs due to Assumption~\ref{ass:Alpha_constant}.
Therefore, the complexity of evaluating the predictive distribution 
scales negligibly with the number of outputs.
This is a major advantage of NNs with BLL,
in comparison to GPs, where it is common to fit an independent model for each output.

\section{Bayes by Backprop}\label{sec:BayesByBackprop}
As a comparative baseline to our proposed method,
we also employ
variational inference with the 
\emph{Bayes by Backprop}~\cite{blundell_Weight_2015, jospinHandsonBayesianNeural2022} method to train a full BNN,
that is, a neural network in which all weights follow a probability distribution. 

\subsection{Background}
For the formulation of the NN with BLL we require the posterior distribution
of the weights in the last layer in~\eqref{eq:Posterior_Bayesian_Linear_Regression}.
As a main difference, we now state the posterior distribution for all weights
of the NN, that is:
\begin{equation}\label{eq:Posterior_BNN}
	p(\sW_{L+1}|\gD, \sTheta)=\frac{
		p(\gD| \sW_{L+1}, \sTheta)p(\sW_{L+1}|\sTheta)
	}
	{
		p(\gD| \sTheta)
	}.
\end{equation}
In contrast to NNs with BLL, 
the exact posterior distribution is intractable 
and we resort to variational inference.
To this end, we introduce a surrogate distribution $q(\sW_{L+1})$
and minimize the Kullback-Leibler (KL) divergence between the surrogate and the true posterior:
\begin{equation}\label{eq:KLD_surrogate_posterior}
	\begin{gathered}
		\DKL(q(\sW_{L+1}) \parallel p(\sW_{L+1}|\gD,\sTheta))\\
		=\E_{q(\sW_{L+1})}
		\left[
		\log\frac{q(\sW_{L+1})}{p(\sW_{L+1}|\gD,\sTheta)}
		\right].    
	\end{gathered}
\end{equation}
Minimizing the KL-divergence yields the surrogate distribution of the weights that 
approximates the true posterior distribution.
As proposed in~\cite{blundell_Weight_2015}, we choose the surrogate distribution as:
\begin{equation}
	q(\vec(\mW_l))  = \N(\vmu_{w,l}, \diag(\vsigma_{w,l}^2)) \quad \forall l \in\sI_{[1,L]},
\end{equation}
with trainable parameters $\vmu_{w,l}$ and $\vsigma_{w,l}$.
Consequently, each weight of the neural network is now represented with two parameters, 
which makes 
Bayes by Backprop a tractable method even for larger models. 

For the concrete implementation of training the BNN with Bayes by Backprop,
we refer to~\cite{jospinHandsonBayesianNeural2022}.
In particular, we use Bayes by Backprop in combination with empirical Bayes 
to obtain the optimal parameters $\sTheta=\{\vsigma_{w},\vsigma_{e}\}$ 
which, similarly to the NN with BLL, include the prior variances of the weights in the last layer,
that is, $\vsigma_{w}$ and the noise covariances $\vsigma_{e}$.
Following the discussion in~\cite{blundell_Weight_2015}, 
we assume a fixed prior for the weights in the previous layers.

\subsection{Evaluation}
The predictive distribution of the trained BNN can be evaluated as:
\begin{equation}\label{eq:BNN_predictive_distribution_exact}
	p(\rvt|\gD,\sTheta, \vx) = \E_{q(\sW_{L+1})}\left[p(\rvt|\vx, \sW_{L+1})\right].
\end{equation}
In contrast to predictive distribution of the NN with BLL in~\eqref{eq:NN_BLL_distribution_targets},
expression~\eqref{eq:BNN_predictive_distribution_exact}
has no analytical solution.
Instead, we resort to Monte Carlo sampling of the weights from the surrogate distribution
and obtain the predictive distribution by averaging over the samples, that is:
\begin{equation}\label{eq:BNN_predictive_distribution_approx}
	\begin{aligned}
		p(\rvt|\gD,\sTheta, \vx) &\approx \frac{1}{N}\sum_{i=1}^{N} p(\rvt|\vx, \sW^{(i)}_{L+1})\\
		&=\frac{1}{N}\sum_{i=1}^{N}\N(\bar\vy^{(i)}, \mSigma_E),
	\end{aligned}
\end{equation}
with $\sW^{(i)}_{L+1}\sim q(\sW_{L+1})$.
Equation~\eqref{eq:BNN_predictive_distribution_approx} 
yields a Gaussian mixture model and also allows for the computation of 
the log-predictive density, similarly to~\eqref{eq:Mean_predictive_probability}.

\section{Simulation study}\noindent\label{ssec:Multivariate_toy_example}
\begin{figure}
	\centering
	\includegraphics[width=\columnwidth]{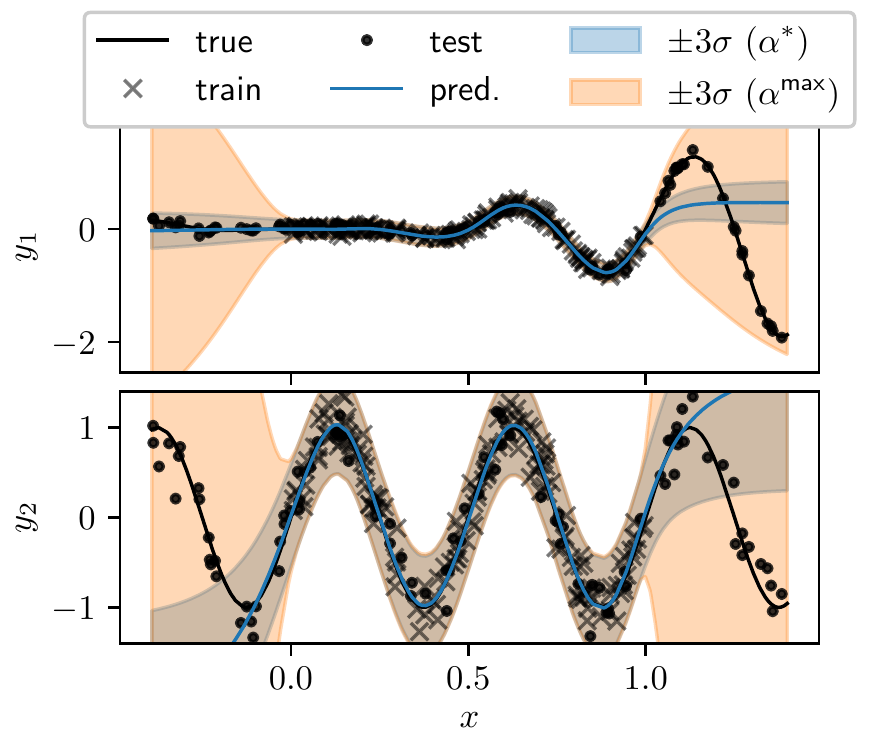}
	\caption{
		Multivariate neural network with Bayesian last layer: Predicted mean and standard deviation for two outputs with different and unknown noise level.
		Training with the proposed Algorithm~\ref{alg:BLL_and_alpha},
		which maximizes the log-marginal likelihood in~\eqref{eq:LMLH_multivariate_simplified},
		and yields the optimal parameter $\alpha^*$.
		This value can then be adapted 
		to improve the uncertainty quantification in the extrapolation regime
		by maximizing~\eqref{eq:Mean_predictive_probability},
		yielding $\alpha^{\max}$.
	}
	\label{fig:07_Multivariate_LMLH}
\end{figure}\noindent
\begin{table}
	\centering
	\renewcommand{\arraystretch}{1.2}
	\caption{Simulation study in Figure~\ref{fig:07_Multivariate_LMLH}: Mean-squared error (MSE), 
		negative LML (NLML) and mean log-predictive density (LPD). Comparison between NN with BLL (trained by minimizing the NLML) 
		and BLR with NN features (NN trained by minimizing MSE).
		Comparison of $\alpha^*$ (maximizing the LML)  vs. $\alpha^{\text{max}}$ (maximizing the LPD for validation data).}
	\label{tab:KPI_Toy_example}
	% \begin{tabularx}{\textwidth}{p{2.5cm}XXXXXXXX}
% {} & \multicolumn{4}{l}{NLML ($\downarrow$ is better)} & \multicolumn{4}{l}{LPD ($\uparrow$ is better)}\\
% \cmidrule(lr){2-5} \cmidrule(lr){6-9}
% {} & \multicolumn{2}{l}{$\alpha^*$} & \multicolumn{2}{l}{$\alpha^{\text{max}}$} & \multicolumn{2}{l}{$\alpha^*$} & \multicolumn{2}{l}{$\alpha^{\text{max}}$} \\
% \cmidrule(lr){2-3} \cmidrule(lr){4-5} \cmidrule(lr){6-7} \cmidrule(lr){8-9} 
% {} &      train &  test &     train &  test &      train &   test &     train &  \textbf{test}\\
% \midrule
% NN w. BLL &      -0.09 & 82.50 &      0.27 & 83.16 &       0.42 & -18.14 &      0.41 & \textbf{-1.83}\\
% BLR w. NN features &       0.14 & 61.06 &      0.62 & 61.90 &       0.29 & -19.20 &      0.29 & \textbf{-1.95}\\
% \bottomrule
% \end{tabularx}
\begin{tabularx}{\columnwidth}{p{2.5cm}XXX}
& &  \multicolumn{2}{l}{LPD ($\uparrow$ is better)}\\
\cmidrule(lr){3-4}
& & train & test\\
\midrule
NN w. BLL & ($\alpha^*$) &0.42 & -18.14 \\
\textbf{NN w. BLL} & ($\alpha^{\text{max}}$) &0.41 & \textbf{-1.83} \\
BLR w. NN features & ($\alpha^*$) &0.29 & -19.20 \\
BLR w. NN features & ($\alpha^{\text{max}}$) &0.29 & -1.95 \\
BNN w. VI~\cite{blundell_Weight_2015} & & 0.00 & -2.30 \\
\bottomrule
\end{tabularx}
\end{table}\noindent
In this section, we investigate our proposed method
in comparison to BLR with features from a trained NN, 
and a full BNN trained with Bayes by Backprop.
The code to reproduce the 
presented results is available online\footnote{\url{https://github.com/4flixt/2023_Paper_BLL_LML}}.

% To this end, we present the simulation study with summarized setting and results shown in Figure~\ref{fig:07_Multivariate_LMLH}.

We investigate data  
from a function with $n_x=1$ input and $n_y=2$ outputs, 
which is shown in Figure~\ref{fig:07_Multivariate_LMLH}.
Both outputs exhibit non-linear behavior with different noise-levels, in particular
$\sigma_1=0.05$ and $\sigma_2=0.2$.
Both noise variances are assumed to be unknown.

We train a NN with BLL using our proposed Algorithm~\ref{alg:BLL_and_alpha}.
%, using the multivariate LML from Result~\ref{res:LMLH_multivariate_simplified}. 
For the study, a suitable structure with $L=2$, $n_{\tilde\phi_1}=n_{\tilde\phi_2}=n_{\tilde\phi_3}=20$ 
and $g_1(\cdot)=g_2(\cdot)=g_3(\cdot)=\tanh(\cdot)$ 
is determined using trial and error. 
To maximize the LML, 
we use the Adam \cite{kingmaAdamMethodStochastic2014} optimizer 
and employ early-stopping to avoid overfitting.
% Finally, we compute the standard deviation of the predicted target distribution according to~\eqref{eq:NN_BLL_distribution_targets}.
% This distribution is visually easier to interpret, as it should contain the noise disturbed train and test samples with a high probability.

The results of the simulation study are shown in Figure~\ref{fig:07_Multivariate_LMLH}.
We visualize mean and standard deviation of the predicted distribution obtained 
from the NN with BLL.
A focus of the investigation is the comparison 
of the distributions 
obtained with 
optimal value $\alpha^*$ and $\alpha^{\text{max}}$.
Both distributions suitably describe the training data 
and capture the behavior of the unknown function in the interpolation regime.
The estimated noise variance 
$\sigma_1=0.051$ and $\sigma_2=0.17$
is close to the true value. 
However, 
only the distribution with tuned extrapolative uncertainty, that is, with $\alpha^{\text{max}}$,
is suitable to describe the extrapolation regime. 
To quantify the improvement, we compute
the log-predictive density 
of the NN with BLL 
for $\alpha^*$ and $\alpha^{\text{max}}$,
and present the results
in Table~\ref{tab:KPI_Toy_example}.
For comparison,
we train a NN with the same structure
by minimizing the mean-squared-error 
and then use the learned features for BLR as in previous works  \cite{mckinnonMetaLearningPaired2021, snoekScalableBayesianOptimization2015a}.
The second step of Algorithm~\ref{alg:BLL_and_alpha}, that is, updating $\alpha$ subsequently with validation data,  
can also be applied in this setting 
and the resulting performance metric is also shown in Table~\ref{tab:KPI_Toy_example}.

The results in Table~\ref{tab:KPI_Toy_example} show that
updating~$\alpha$, as proposed in Algorithm~\ref{alg:BLL_and_alpha},
significantly improves the log-predictive density~\eqref{eq:Mean_predictive_probability}
of the test data,
both for BLL and BLR.
We see that the NN with BLL, which is trained according to Algorithm~\ref{alg:BLL_and_alpha},
achieves 
overall the best results with respect to the log-predictive density.

\subsection{Comparison to Bayes by Backprop}\label{ssec:Comparison_BayesByBackprop}\noindent
As a final step of our evaluation, we compare the NN with BLL to a full BNN trained with Bayes by Backprop,
as described in Section~\ref{sec:BayesByBackprop}.
We investigate the same neural network architecture 
as described in the previous subsection
and, as a main difference,
now have distributed weights in all layers.
We determine a fixed prior distribution for the weights in all but the last layer as:
\begin{equation}
	p(\vec(\mW_{l,0}))  = \N(0, 0.5 \mI) \quad \forall l \in\sI_{[1,L]}.
\end{equation}
Setting the mean of the prior distribution to zero is justified by employing 
batch normalization \cite{ioffe_Batch_2015} after each variational layer.
As for the NN with BLL, the prior variance of the weights in the last layer is also learned.

\begin{figure}
	\centering
	\includegraphics[width=\columnwidth]{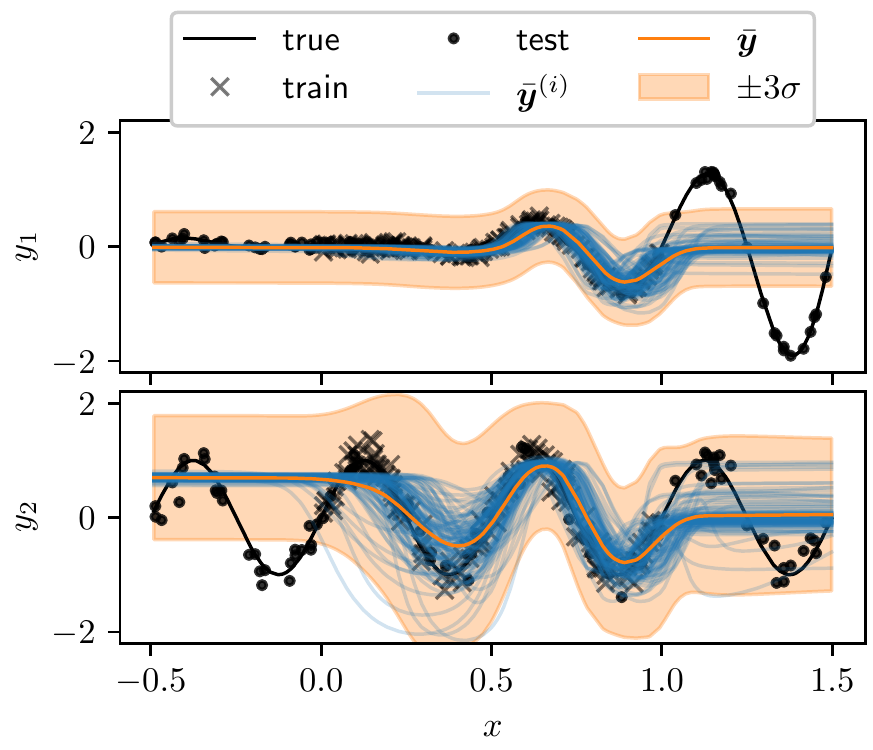}
	\caption{
		Variational inference for a BNN with Bayes by Backprop. 
		Sampled ($N=100$) predictive distribution with~\eqref{eq:BNN_predictive_distribution_approx}.
	}
	\label{fig:bnn_vi_toy_example}
\end{figure}\noindent

Training the BNN with Bayes by Backprop is performed as outlined in Section~\ref{sec:BayesByBackprop}
and using the Adam optimizer. %with a learning rate of $0.01$ for 10.000 epochs. 
The resulting predictive distribution is obtained by sampling $N=100$ weights from the surrogate distribution
and displayed in Figure~\ref{fig:bnn_vi_toy_example}.
We show the mean values $\bar\vy^{(i)}$ 
that are obtained with the sampled weights $\sW^{(i)}_{L+1}\sim q(\sW_{L+1})$.
Additionally, we compute mean and variance of the resulting 
Gaussian mixture model in~\eqref{eq:BNN_predictive_distribution_approx}
and display them in the same figure. 
Finally,
we compute the log-predictive density of the Gaussian mixture model 
and report the results in Table~\ref{tab:KPI_Toy_example}. 
We find that the BNN trained with Bayes by Backprop yields a suitable predictive distribution
but is outperformed by the NN with BLL.
In particular, we see that the extrapolative uncertainty of the NN with BLL 
is better suited to describe the data. 
This holds especially after tuning $\alpha$ with the validation data,
as described in Algorithm~\ref{alg:BLL_and_alpha}.
For a BNN trained with Bayes by Backprop 
it is not possible to modify~$\alpha$ 
to calibrate the extrapolation behavior of the model.
Furthermore,
it is necessary to sample weights from the surrogate posterior
to obtain the predictive distribution, 
adding significant complexity to this method.
This is in contrast to the NN with BLL, 
where the predictive distribution can be computed analytically.

\section{Conclusions}\label{sec:Conclusions}\noindent
Neural networks with Bayesian last layer 
are an attractive compromise between tractability and expressiveness 
in the field of Bayesian neural networks.
As a main contribution of this work, we 
propose 
an efficient algorithm for training neural networks
with Bayesian last layer, based on 
a reformulation of the marginal likelihood.
Importantly, we show that our reformulation satisfies the conditions of optimality
for the same argument as the original formulation.
In comparison to Bayesian linear regression with features from a previously trained neural network,
we find that training on the log-marginal likelihood 
shows advantages
in a presented simulation study.
Our second contribution, an algorithm to tune the extrapolative uncertainty
also shows excellent results in the simulation study.
To derive this algorithm, we further contribute a discussion on 
the relationship of the Bayesian last layer covariance and a proposed
metric to quantify extrapolation.
This metric intuitively relates to the definition of extrapolation 
which is based on the convex hull.
Our proposed method also compares favorably to 
a full Bayesian neural network trained with Bayes by Backprop.
Important advantages are 
the possibility to tune the extrapolative uncertainty,
an analytical expression for the predictive distribution
and the overall ease of implementation. 

This work lays the theoretical foundations for our future work:
identification of probabilistic dynamic system models 
that can be used for safe control decisions.

\bibliography{2023_UQ_NN_BLL_literature.bib}

\begin{thebibliography}{10}
\providecommand{\url}[1]{#1}
\csname url@samestyle\endcsname
\providecommand{\newblock}{\relax}
\providecommand{\bibinfo}[2]{#2}
\providecommand{\BIBentrySTDinterwordspacing}{\spaceskip=0pt\relax}
\providecommand{\BIBentryALTinterwordstretchfactor}{4}
\providecommand{\BIBentryALTinterwordspacing}{\spaceskip=\fontdimen2\font plus
\BIBentryALTinterwordstretchfactor\fontdimen3\font minus
  \fontdimen4\font\relax}
\providecommand{\BIBforeignlanguage}[2]{{%
\expandafter\ifx\csname l@#1\endcsname\relax
\typeout{** WARNING: IEEEtran.bst: No hyphenation pattern has been}%
\typeout{** loaded for the language `#1'. Using the pattern for}%
\typeout{** the default language instead.}%
\else
\language=\csname l@#1\endcsname
\fi
#2}}
\providecommand{\BIBdecl}{\relax}
\BIBdecl

\bibitem{jospinHandsonBayesianNeural2022}
L.~V. Jospin, H.~Laga, F.~Boussaid, W.~Buntine, and M.~Bennamoun, ``Hands-on
  {{Bayesian}} neural networks - {{A}} tutorial for deep learning users,''
  \emph{IEEE Computational Intelligence Magazine}, vol.~17, no.~2, pp. 29--48,
  2022.

\bibitem{abdar_review_2021}
M.~Abdar, F.~Pourpanah, S.~Hussain, D.~Rezazadegan, L.~Liu, M.~Ghavamzadeh,
  P.~Fieguth, X.~Cao, A.~Khosravi, and U.~R. Acharya, ``A review of uncertainty
  quantification in deep learning: {{Techniques}}, applications and
  challenges,'' \emph{Information Fusion}, vol.~76, pp. 243--297, 2021.

\bibitem{harrisonControlAdaptationMetalearning2018}
J.~Harrison, A.~Sharma, R.~Calandra, and M.~Pavone, ``Control adaptation via
  meta-learning dynamics,'' in \emph{Workshop on {{Meta-Learning}} at
  {{NeurIPS}}}, 2018.

\bibitem{mckinnonMetaLearningPaired2021}
C.~D. McKinnon and A.~P. Schoellig, ``Meta learning with paired forward and
  inverse models for efficient receding horizon control,'' \emph{IEEE Robotics
  and Automation Letters}, vol.~6, no.~2, pp. 3240--3247, 2021.

\bibitem{wabersichNonlinearLearningbasedModel2021}
K.~P. Wabersich and M.~N. Zeilinger, ``Nonlinear learning-based model
  predictive control supporting state and input dependent model uncertainty
  estimates,'' \emph{International Journal of Robust and Nonlinear Control},
  vol.~31, no.~18, pp. 8897--8915, 2021.

\bibitem{dongRobustDataDrivenIterative2021}
J.~Dong, ``Robust {{Data-Driven Iterative Learning Control}} for
  {{Linear-Time-Invariant}} and {{Hammerstein-Wiener Systems}},'' \emph{IEEE
  Transactions on Cybernetics}, pp. 1--14, 2021.

\bibitem{mckinnonLearningProbabilisticModels2019}
C.~D. McKinnon and A.~P. Schoellig, ``Learning {{Probabilistic Models}} for
  {{Safe Predictive Control}} in {{Unknown Environments}},'' in
  \emph{Proceedings of the 18th {{European Control Conference}}}, 2019, pp.
  2472--2479.

\bibitem{hewingCautiousModelPredictive2020}
L.~Hewing, J.~Kabzan, and M.~N. Zeilinger, ``Cautious {{Model Predictive
  Control Using Gaussian Process Regression}},'' \emph{IEEE Transactions on
  Control Systems Technology}, vol.~28, no.~6, pp. 2736--2743, 2020.

\bibitem{rasmussenGaussianProcessesMachine2003}
C.~E. Rasmussen, \emph{Gaussian Processes in Machine Learning}.\hskip 1em plus
  0.5em minus 0.4em\relax {Springer}, 2003.

\bibitem{lazaro-gredillaMarginalizedNeuralNetwork2010}
M.~{L{\'a}zaro-Gredilla} and A.~R. {Figueiras-Vidal}, ``Marginalized neural
  network mixtures for large-scale regression,'' \emph{IEEE transactions on
  neural networks}, vol.~21, no.~8, pp. 1345--1351, 2010.

\bibitem{liuWhenGaussianProcess2020}
H.~Liu, Y.-S. Ong, X.~Shen, and J.~Cai, ``When {{Gaussian Process Meets Big
  Data}}: {{A Review}} of {{Scalable GPs}},'' \emph{IEEE Transactions on Neural
  Networks and Learning Systems}, vol.~31, no.~11, pp. 4405--4423, 2020.

\bibitem{wilsonDeepKernelLearning2016}
A.~G. Wilson, Z.~Hu, R.~Salakhutdinov, and E.~P. Xing, ``Deep {{Kernel
  Learning}},'' in \emph{Proceedings of the 19th {{International Conference}}
  on {{Artificial Intelligence}} and {{Statistics}}}.\hskip 1em plus 0.5em
  minus 0.4em\relax {PMLR}, 2016, pp. 370--378.

\bibitem{liuDeepLatentVariableKernel2022}
H.~Liu, Y.-S. Ong, X.~Jiang, and X.~Wang, ``Deep {{Latent-Variable Kernel
  Learning}},'' \emph{IEEE Transactions on Cybernetics}, vol.~52, no.~10, pp.
  10\,276--10\,289, 2022.

\bibitem{Lecun2015}
Y.~Lecun, Y.~Bengio, and G.~Hinton, ``Deep learning,'' \emph{Nature}, vol. 521,
  no. 7553, pp. 436--444, 2015.

\bibitem{karg_Efficient_2020}
B.~Karg and S.~Lucia, ``Efficient representation and approximation of model
  predictive control laws via deep learning,'' \emph{IEEE Transactions on
  Cybernetics}, vol.~50, no.~9, pp. 3866--3878, 2020.

\bibitem{dongFunctionalNonlinearModel2019}
L.~Dong, J.~Yan, X.~Yuan, H.~He, and C.~Sun, ``Functional {{Nonlinear Model
  Predictive Control Based}} on {{Adaptive Dynamic Programming}},'' \emph{IEEE
  Transactions on Cybernetics}, vol.~49, no.~12, pp. 4206--4218, 2019.

\bibitem{sarangapaniSystemIdentificationUsing2006}
J.~Sarangapani, ``System {{Identification Using Discrete-Time Neural
  Networks}},'' in \emph{Neural {{Network Control}} of {{Nonlinear
  Discrete-Time Systems}}}, 1st~ed.\hskip 1em plus 0.5em minus 0.4em\relax {CRC
  Press}, 2006, pp. 443--466.

\bibitem{fiedlerEconomicNonlinearPredictive2020}
F.~Fiedler, A.~Cominola, and S.~Lucia, ``Economic nonlinear predictive control
  of water distribution networks based on surrogate modeling and automatic
  clustering,'' \emph{IFAC-PapersOnLine}, vol.~53, no.~2, pp. 16\,636--16\,643,
  2020.

\bibitem{fiedlerModelPredictiveControl2022}
F.~Fiedler and S.~Lucia, ``Model predictive control with neural network system
  model and {{Bayesian}} last layer trust regions,'' in \emph{Proceedings of
  the 17th {{IEEE International Conference}} on {{Control}} \& {{Automation}}},
  2022, pp. 141--147.

\bibitem{salakhutdinovBayesianProbabilisticMatrix2008}
R.~Salakhutdinov and A.~Mnih, ``Bayesian probabilistic matrix factorization
  using {{Markov}} chain {{Monte Carlo}},'' in \emph{Proceedings of the 25th
  International Conference on {{Machine}} Learning}, 2008, pp. 880--887.

\bibitem{blundell_Weight_2015}
C.~Blundell, J.~Cornebise, K.~Kavukcuoglu, and D.~Wierstra, ``Weight
  uncertainty in neural network,'' in \emph{Proceedings of the International
  Conference on Machine Learning}.\hskip 1em plus 0.5em minus 0.4em\relax
  {PMLR}, 2015, pp. 1613--1622.

\bibitem{watsonLatentDerivativeBayesian2021}
J.~Watson, J.~A. Lin, P.~Klink, J.~Pajarinen, and J.~Peters, ``Latent
  {{Derivative Bayesian Last Layer Networks}},'' in \emph{Proceedings of the
  International {{Conference}} on {{Artificial Intelligence}} and
  {{Statistics}}}.\hskip 1em plus 0.5em minus 0.4em\relax {PMLR}, 2021, pp.
  1198--1206.

\bibitem{bishopPatternRecognitionMachine2006}
C.~M. Bishop, \emph{Pattern Recognition and Machine Learning}.\hskip 1em plus
  0.5em minus 0.4em\relax {Springer}, 2006.

\bibitem{snoekScalableBayesianOptimization2015a}
J.~Snoek, O.~Rippel, K.~Swersky, R.~Kiros, N.~Satish, N.~Sundaram, M.~Patwary,
  M.~Prabhat, and R.~Adams, ``Scalable bayesian optimization using deep neural
  networks,'' in \emph{Proceedings of the International Conference on Machine
  Learning}.\hskip 1em plus 0.5em minus 0.4em\relax {PMLR}, 2015, pp.
  2171--2180.

\bibitem{balestrieroLearningHighDimension2021}
R.~Balestriero, J.~Pesenti, and Y.~LeCun, ``Learning in {{High Dimension Always
  Amounts}} to {{Extrapolation}},'' \emph{arXiv:2110.09485 [cs]}, 2021.

\bibitem{bernardoBayesianTheory1994a}
J.~M. Bernardo and A.~F. Smith, \emph{Bayesian Theory}.\hskip 1em plus 0.5em
  minus 0.4em\relax {John Wiley \& Sons}, 1994, vol. 405.

\bibitem{fazelLogdetHeuristicMatrix2003}
M.~Fazel, H.~Hindi, and S.~Boyd, ``Log-det heuristic for matrix rank
  minimization with applications to {{Hankel}} and {{Euclidean}} distance
  matrices,'' in \emph{Proceedings of the {{American Control Conference}}},
  vol.~3, 2003, pp. 2156--2162 vol.3.

\bibitem{dongCompressiveSensingNonlocal2014}
W.~Dong, G.~Shi, X.~Li, Y.~Ma, and F.~Huang, ``Compressive {{Sensing}} via
  {{Nonlocal Low-Rank Regularization}},'' \emph{IEEE Transactions on Image
  Processing}, vol.~23, no.~8, pp. 3618--3632, 2014.

\bibitem{kerstingMostLikelyHeteroscedastic2007}
K.~Kersting, C.~Plagemann, P.~Pfaff, and W.~Burgard, ``Most likely
  heteroscedastic {{Gaussian}} process regression,'' in \emph{Proceedings of
  the 24th International Conference on {{Machine}} Learning}.\hskip 1em plus
  0.5em minus 0.4em\relax {Association for Computing Machinery}, 2007, pp.
  393--400.

\bibitem{seber_matrix_2008}
G.~A.~F. Seber, \emph{A Matrix Handbook for Statisticians}.\hskip 1em plus
  0.5em minus 0.4em\relax {John Wiley \& Sons}, 2008.

\bibitem{kingmaAdamMethodStochastic2014}
D.~P. Kingma and J.~Ba, ``Adam: {{A}} method for stochastic optimization,''
  \emph{arXiv preprint arXiv:1412.6980}, 2014.

\bibitem{ioffe_Batch_2015}
S.~Ioffe and C.~Szegedy, ``Batch {{Normalization}}: {{Accelerating Deep Network
  Training}} by {{Reducing Internal Covariate Shift}},'' in \emph{Proceedings
  of the 32nd {{International Conference}} on {{Machine Learning}}}.\hskip 1em
  plus 0.5em minus 0.4em\relax {PMLR}, 2015, pp. 448--456.

\end{thebibliography}

\balance

\clearpage

\end{document}